\documentclass[lettersize,journal]{IEEEtran}
\usepackage{amsmath,amsfonts}
\usepackage{algorithmic}
\usepackage{algorithm}
\usepackage{array}
\usepackage[caption=false,font=normalsize,labelfont=sf,textfont=sf]{subfig}
\usepackage{textcomp}
\usepackage{stfloats}
\usepackage{url}
\usepackage{verbatim}
\usepackage{graphicx}
\usepackage{cite}
\usepackage{soul}
\usepackage{booktabs}
\usepackage{xcolor}
\usepackage{bbding}
\usepackage[misc]{ifsym}

\usepackage[switch]{lineno}

\begin{document}

\title{LCPFormer: Towards Effective 3D Point Cloud Analysis via Local Context Propagation in Transformers}

\author{Zhuoxu~Huang*,
        Zhiyou~Zhao*,
        Banghuai~Li \textsuperscript{\Envelope},
        and~Jungong~Han,~\IEEEmembership{Member,~IEEE,}
\thanks{Zhuoxu Huang is with the Department of Computer Science, Aberystwyth University, Aberystwyth SY23 3DB, U.K. (e-mail: zhh6@aber.ac.uk). Zhuoxu Huang is also with AnMai Technology (Ningbo) Co., Ltd.}
\thanks{Zhiyou Zhao is with the Department of Computer Science and Engineering, Shanghai Jiao Tong University, Shanghai 200240 China. (e-mail: zhaozhiyou789@sjtu.edu.cn).}
\thanks{Banghuai Li is with the School of Electronics Engineering and Computer Science, Peking University,  Beijing 100871 China. (e-mail: libanghuai@pku.edu.cn).}
\thanks{Jungong Han is with the Department of Computer Science, University of Sheffield, Sheffield S10 2TN, U.K. (e-mail: jungonghan77@gmail.com).}
\thanks{* Authors have equal contributions in this work.}
\thanks{\textsuperscript{\Envelope} Corresponding author: libanghuai@pku.edu.cn}
\thanks{Manuscript received April 19, 2021; revised August 16, 2021.}}

\markboth{Journal of \LaTeX\ Class Files,~Vol.~14, No.~8, August~2021}%
{Shell \MakeLowercase{\textit{et al.}}: A Sample Article Using IEEEtran.cls for IEEE Journals}

\IEEEpubid{0000--0000/00\$00.00~\copyright~2023 IEEE}

\maketitle

\begin{abstract}
Transformer with its underlying attention mechanism and the ability to capture long-range dependencies makes it become a natural choice for unordered point cloud data. However, local regions separated from the general sampling architecture corrupt the structural information of the instances, and the inherent relationships between adjacent local regions lack exploration. In other words, the transformer only focuses on the long-range dependence, while local structural information is still crucial in a transformer-based 3D point cloud model. To enable transformers to incorporate local structural information, we proposed a straightforward solution based on the natural structure of the point clouds to exploit the message passing between neighboring local regions, thus making their representations more comprehensive and discriminative. Concretely, the proposed module, named \textbf{Local Context Propagation (LCP)}, is inserted between two transformer layers. It takes advantage of the overlapping points of adjacent local regions (statistically shown to be prevalent) as intermediaries, then re-weighs the features of these shared points from different local regions before passing them to the next layers. Finally, we design a flexible LCPFormer architecture equipped with the LCP module, which is applicable to several different tasks. Experimental results demonstrate that our proposed LCPFormer outperforms various transformer-based methods in benchmarks including 3D shape classification and dense prediction tasks such as 3D object detection and semantic segmentation. Code will be released for reproduction.
\end{abstract}


\section{Introduction}
\label{sec:intro}
\IEEEPARstart{I}{ncreasing} real-world applications, including robotics, autonomous driving, and augmented reality, enhanced the use of 3D point cloud data due to its precise distance perception and information capacity. However, unlike 2D images in the form of regular pixel grids, it is challenging to effectively ingest features from 3D point clouds due to their sparsity and unstructured nature.

\begin{figure}[htbp]
\centering
\includegraphics[width=1.0\linewidth]{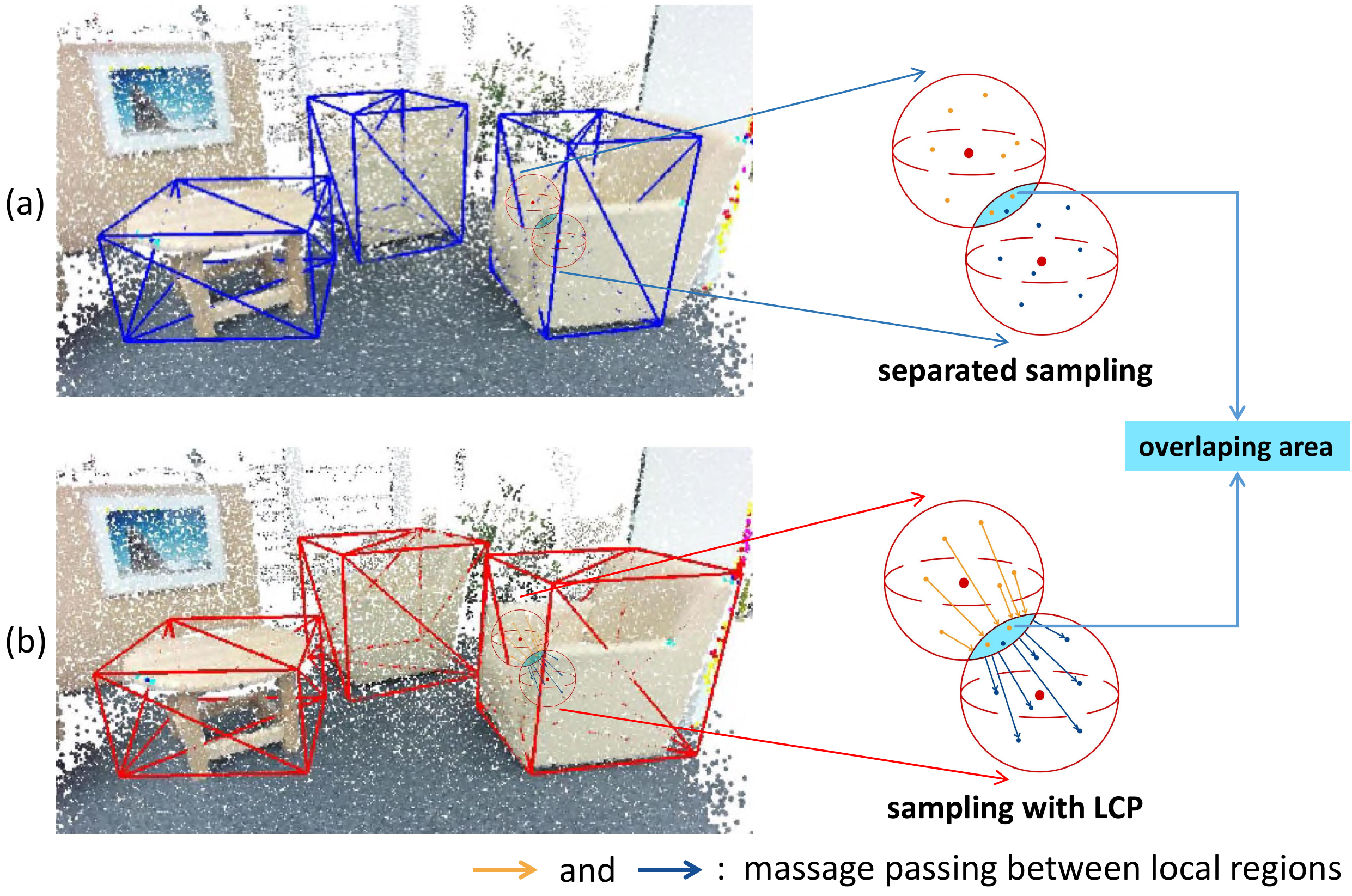}
\caption{ (a) Separate local regions that lack whole structure information lead to inaccurate detection boxes (\textcolor{blue}{blue boxes}). (b) Using overlapping points as intermediaries, our LCP module can pass information between neighboring local regions (the left part) and produce more precise detection results (\textcolor{red}{red boxes}).}
\label{motivation}
\end{figure}

Based on how point clouds are represented, existing feature extraction methods can roughly be divided into three categories: projection-based, voxel-based and point-based methods. Projection-based methods \cite{deng2021h2rcnn,Liu2021PQANet,yang2018pixor,liang2018deep,ku2018joint} first project 3D point clouds into a 2D plane and then treat them as pixel representations for processing. Voxel-based methods \cite{yan2018second,song2016deep,SparseConv2018graham,Minkowski,lang2019pointpillars,3dsis,pillar,zhou2018voxelnet} usually voxelize 3D point clouds into uniform grids and then apply 3D convolutional operations on the resulting volumetric representations. These methods provide a fair solution for applying convolution on unstructured 3D point clouds. However, voxel-based methods cannot scale well to sparse 3D data because of the cubical computation cost with the resolution and both projection/voxel-based approaches suffer from inevitable structural information loss resulting from the quantization process. 
In contrast, point-based methods \cite{qi2017pointnet,qi2017pointnet++,thomas2019kpconv,hu2020randla,qi2019votenet,yogo} directly operate on raw point clouds to learn 3D representations. They do not introduce explicit information loss resulting from the voxelization or projection process and become increasingly popular. But using simple multi-layer perceptions (MLP) as the basic operator limits the learning capability of the network.

\IEEEpubidadjcol
Transformers\cite{guo2021pct,pan2021pointformer,zhao2021pointtransformer,engel2021pointtransformer,liu2021groupfree,wang2021spatial} are also adopted recently as point-based methods in 3D point cloud processing after achieving a great success in solving various 2D vision tasks \cite{zhu2020deformableDETR,carion2020DETR,dosovitskiy2020ViT,chen2021crossvit,d2021convit,yuan2021hrformer,han2021transformer}. On the one hand, the transformer with its underlying multi-head self-attention mechanism owns the property of permutation invariance which is consistent with the characteristics of point clouds. On the other hand, it is highly expressive with the ability to capture long-range dependencies and effectively learn context-aware representations. They both make the transformer a natural fit for 3D point clouds. However, the absence of local information integration has been described as an inherent deficiency of transformer networks \cite{liu2021swin}. While several methods have implemented different transformer-based models for 3D point clouds, the local structural information is often disregarded. PCT \cite{guo2021pct} sends point clouds through an embedding layer and directly applies attention to the entire set of points to extract features. However, this implies the demands on heavy memory and computation overheads. To reduce the complexity, Point Transformer \cite{zhao2021pointtransformer} first groups point clouds into several local regions, followed by attention operations in each local region independently. Consequently, each point can receive information from every other point in the same local region, thus capturing local structural patterns. However, it still faces the problem arising from the separate local regions sampling, which leads to the destruction of the instance structure since the inherent relationship between neighboring local regions lacks exploration.
Although \cite{pan2021pointformer} goes a step further to apply max-pooling on each local region and perform the self-attention operation again on those pooled features (named global transformer layer \cite{pan2021pointformer}), it still encounters the same problem because the extra self-attention operation only focuses on modeling the scene-level dependencies.

The aforementioned problems have been studied in 2D scenes, and different solutions were proposed to transmit information amongst local windows \cite{liu2021swin,yuan2021hrformer}. However, unlike 2D images, 3D point clouds are irregular, unordered, and unstructured representations. The interactions between local regions are also irregular in unstructured positions instead of regular grid locations. Therefore, it is much more difficult in 3D scenes to model the interactions between local regions on the fine-grained point level. To solve this problem, we take an in-depth analysis of transformers in 3D point clouds and propose to promote message passing among those neighboring local regions that share the same points. In this way, each local region can obtain more informative and discriminative features to boost the ability of transformers. Based on this thought, we propose an effective yet lightweight module, named {\bf Local Context Propagation (LCP)}. 

The LCP module is inserted between two transformer layers such that the overall framework looks like a sandwich structure. Specifically, when following the standard settings in \cite{qi2017pointnet++,pan2021pointformer} to adopt Farthest Point Sampling (FPS) and k-Nearest Neighbors (kNN) to divide the whole point cloud into different local regions, we find statistically that most local regions tend to overlap with others. Based on this natural structure of the point clouds, as shown in Fig.\ref{motivation}(b), those \textit{shared} points in overlapping areas of the different local regions can serve as the \textbf{carrier} to perform message passing among neighboring local regions exactly. After getting the point feature from the last transformer layer, we collect all local regions that each \textit{shared} point belongs to, then adopt a lightweight network to generate an adaptive weight acting on those regions. Finally, we update the feature of this point by aggregating its corresponding features from different local regions using the obtained weight. These features updated by the LCP module will then pass to all the other points in a local region with the self-attention mechanism in the next transformer layer. This way allows us to propagate the contexts between neighboring local regions successfully. It is worth noting that the proposed LCP module also differs from the message aggregation that happens between local regions via indirect progressive processing of the larger region with the bigger radius. Our LCP module enables direct message passing between local regions that gives the network a direct learning target and thus enhances the point-level features. In more detail, the main difference between the direct/indirect approach is that our method is applied before max-pooling which is used during every radius change. Thus the message between local regions is interacted before the max-pooling aggregation to minimize information loss when using LCP, which is critical due to the significant global nature of the transformers.

Finally, based on the proposed LCP module, we design a flexible architecture named \textbf{LCPFormer}. Equipped with different numbers of the sandwich structures we mentioned above, which we call \textbf{LCPFormer Block} (shown in Fig.\ref{overallarchi}), the LCPFormer is suitable for different 3D point cloud tasks. In summary, the contributions of this work are three-fold:
\begin{itemize}
\item We delve deep into the transformer-based methods in 3D point cloud analysis and propose a simple yet effective module named LCP to promote message passing between neighboring local regions.
\item Based on our proposed LCP module, a transformer-based backbone architecture is designed and can be applied for various 3D perception tasks, including object detection, semantic segmentation, and shape classification.
\item Our proposed LCPFormer consistently outperforms different transformer-based methods in various 3D tasks including 3D shape classification (93.6\% overall accuracy on ModelNet40) and 3D object detection (63.2\% mAP@0.25 and 46.2\% mAP@0.5 on SUN RGB-D) and semantic segmentation (63.4\% mIoU on SensatUrban; 70.2\% mIoU in S3DIS Area 5). 
\end{itemize}

\section{Related Work}
\label{sec:relatedwork}
\subsection{Representation Learning for 3D Point Clouds.}
Different from 2D images, 3D point clouds with their disorder and sparsity make them extremely difficult to be processed. Previous works mainly include feature extraction by projection, voxelization, directly processing point clouds, and the fusion of those. 

\textbf{Projection-based} methods implement different projections to view the 3D point clouds into regular representations on 2D planes. Then apply the 2D convolution network for the features extraction. Methods including projecting points to the perspective view, \cite{deng2021h2rcnn}, the bird-eye view \cite{yang2018pixor,liang2018deep,ku2018joint}, the tangent plane \cite{TragenConv2018tangent}, and the use of multi-view fusion \cite{mvcnn2015,Liu2021PQANet,qi2016volumetric,kanezaki2018rotationnet}. For example, \cite{deng2021h2rcnn} hallucinates the 3D representation by taking full advantage of complementary information in the perspective view and the bird-eye view. \cite{Liu2021PQANet} adopts a multi-view projection strategy to effectively extract features for point cloud quality assessment. Inevitably, lots of points collapsed thus causing the geometric information loss during the projection. 
In addition, network performance can be severely affected by the fact that the 2D representation differs due to the occlusion in 3D point clouds and different projection methods.

\textbf{Voxel-based} methods \cite{maturana2015voxnet,song2017semantic,zhou2018voxelnet} is another solution that converts 3D points into voxel grids and applies 3D convolution to generate features. However, the 3D convolution suffers from the cubical computation cost with the resolution when applied directly on the voxel. Although some approaches make good use of the sparsity of the point cloud with sparse convolutions \cite{Minkowski,SparseConv2018graham}, they can not combat the loss of structural information during the quantification.

\textbf{Point-based} methods are proposed in order to prevent the loss of geometric information caused by projection or voxelization. PointNet \cite{qi2017pointnet} processes the 3D point clouds directly with point-wise multi-layer perceptions (MLP) and pooling, while PointNet++ \cite{qi2017pointnet++} further designs a hierarchical structure by dividing the whole point cloud into separate local regions using querying and grouping. Various structures are proposed based on such models \cite{hu2020randla,wu2019pointconv,thomas2019kpconv,yang2019gumbelsampling,dovrat2019snet}. For example, \cite{zhao2021transformer3ddet} produces produce more accurate voting centers for object detection by modeling the relationship from neighboring clusters. Alternatively, \cite{Li2022psnet} PSNet proposes a fast data structuring method to tackle the data structuring issue in point-based methods. There are also other methods that combine both point-wise and voxel-wise features in a local field \cite{lang2019pointpillars,guan2021m3detr} or a global field \cite{ye2020hvnet,shi2020pv,pointgrid2018le}. In addition, some of these approaches propose using continuous convolution on the 3D point clouds by the designed kernel structure. For example, PointConv \cite{wu2019pointconv} design the convolution kernels as nonlinear functions related to the distance between point coordinates. KPConv \cite{thomas2019kpconv} further proposes a deformable convolution whose weights are placed in the Euclidean space by the kernel. In order to ingest the in-depth geometric information of the point cloud, some methods \cite{SPGraph2018Landrieu,shen2018kcnet,wang2019graphattn,Li2019deepgcn} propose creating graphs based on the coordinates and geometry of the point clouds and then adapting the graph convolutions on these point graphs. DGCNN \cite{wang2019dgcnn} propose an EdgeConv that can be dynamically updated in layers leveraging the properties of the diagram. PointWeb \cite{pointweb2019zhao} explores the interaction between points by dense connection. Our work focuses on processing point clouds directly via transformers.

\subsection{Transformers in 2D Vision.}
Inspired by the great success of transformers in the natural language processing field, people start to consider applying transformers to vision tasks.
\cite{dosovitskiy2020ViT} divides the image into independent patches, and then reshapes each patch into a vector as the input element of the transformer. Since then, transformer structure has been successfully applied to computer vision, and many improvements\cite{d2021convit,touvron2021going,wu2021cvt,graham2021levit,yuan2021tokens,han2021transformer,chen2021crossvit,liu2021swin,jiang2021all,srinivas2021bottleneck,carion2020DETR,zhu2020deformableDETR} have been proposed. Some methods \cite{liu2021swin,yuan2021hrformer,huang2019interlaced,vaswani2021scaling} further group images with non-overlapping windows and apply transformers within each window separately, but lack feature aggregation between each window. To deal with this problem, \cite{liu2021swin} proposes shifted windows to dynamically adjust the window position and can be used for multiple tasks including semantic segmentation, object detection, and image classification. \cite{yuan2021hrformer} combines transformers and the HRNet \cite{wang2020hrnet}, borrowing the natural advantage of a $3\times3$ depth-wise convolution to improve transformers, \cite{yuan2021hrformer} 
is also compatible with dense prediction tasks such as human pose estimation.

\subsection{Transformers in 3D Point Clouds.}
As a sequence-to-sequence structure, transformers completely ignore the input order and treat the position information as a set of sequences \cite{vaswani2017attention,zhao2020exploring}, revealing the possibility of the application of transformers in the 3D field. Currently, many transformer-based methods for 3D point clouds have been proposed to improve the performance of PointNet++ \cite{qi2017pointnet++}. Point Transformer \cite{zhao2021pointtransformer} proposes a U-Net \cite{ronneberger2015unet} structure with self-attention layers constructed from vector attention \cite{zhao2020exploring}. This model can be used in both shape classification and object part segmentation tasks in 3D point clouds. Similarly, \cite{engel2021pointtransformer} also proposes a similar idea that uses the attention mechanism and designs a SortNet to refine local features from different sub-spaces. PCT \cite{guo2021pct} proposes to use an offset-attention with implicit Laplace operators and normalization refinement for transformer layers. 

Pu-Transformer \cite{qiu2022putransformer} is introduced for point cloud upsampling that uses transformer structure to enhance point-wise and channel-wise relations of the point feature.

For the detection task, Pointformer \cite{pan2021pointformer} designs an attention-based feature extractor to serve as the backbone for different object detection frameworks. \cite{liu2021groupfree} adopts PointNet++ as the backbone and proposes a Group-Free detector with multi-head self-attention.

In this paper, we take an in-depth analysis of the drawbacks of transformer-based approaches in 3D point clouds. We find that separate local regions damage the structure of instances, which restricts the ability of transformers to enhance the feature representations. Thus, we design a simple yet effective method to exploit the inherent relationship between local regions.

\section{Method}
In this section, we first briefly revisit the general formulation of the transformer and self-attention. Then we conduct an in-depth analysis of transformers in 3D point clouds and propose a lightweight yet effective module, named Local Context Propagation Module (LCP). Lastly, based on this module, we present our LCPFormer architecture for various 3D tasks including shape classification, object detection, and semantic segmentation.

\begin{figure*}[htbp]
\centering
\includegraphics[width=1.0\linewidth]{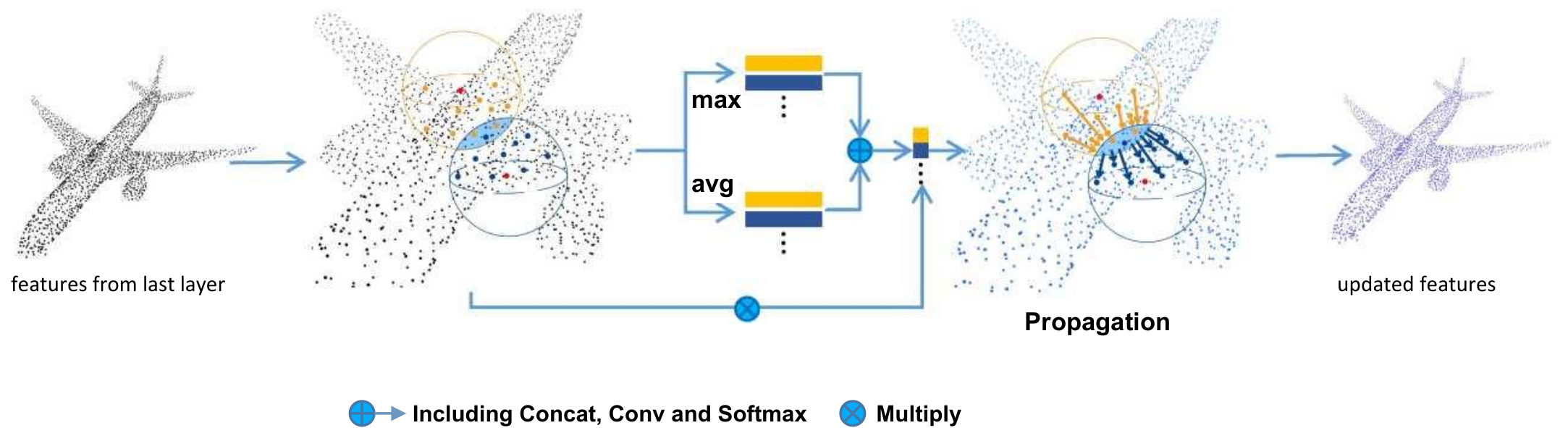}
   \caption{Illustration of the Local Context Propagation module. Given a point shared by several local regions, an adaptive weight acting on those regions is generated. The feature of this point is updated by aggregating its corresponding features from different local regions using the obtained weight. Thus context information is propagated between neighboring local regions effectively. The mean and max operations are for weight generation when updating those overlapping point features.}
\label{cntstructure}
\end{figure*}

\subsection{Preliminary}
\label{preliminary}
We start by reviewing the commonly used multi-head self-attention (MHSA) in transformers. Given a set of input features $F=\{f_i\}$ and corresponding positions $X=\{x_i\}$, the MHSA operation aggregates contents from all input elements according to the computed attention weights. It can be formulated as follows:
\begin{equation}
    F' = \mathrm{PE}(X) + F
\end{equation}
\begin{equation}
    Q_m = F'W_m^{Q},\ K_m = F'W_m^{K},\ V_m = F'W_m^{V}
\end{equation}
\begin{equation}
    H_m = \sigma(Q_m K_m^T / \sqrt{d})V_m
\end{equation}
\begin{equation}
    O = \mathrm{Concat}(H_1,\dots,H_m)W^{O}
\end{equation}
, where $W_m^Q, W_m^K, W_m^V$ are learnable projections of the $m$th head for query, key, and value respectively. $\mathrm{PE}(\cdot)$ is the positional encoding function that projects input coordinates to the same dimension as input features, to which they are later added. $\sigma$ denotes a normalization function $\mathrm{Softmax}$, and $d$ is the feature dimension.

A transformer layer consists of a multi-head self-attention (MHSA), and an element-wise feed-forward network (FFN) with skip connections:

\begin{equation}
    Y = \mathrm{MHSA}(X, F) + F\\
\end{equation}
\begin{equation}
    O = Y + \mathrm{FFN}(Y)
\end{equation}

A vanilla transformer block usually contains several consecutive transformer layers mentioned above.

\subsection{Rethinking Transformers in 3D Point Clouds}
\label{rethinkingtransformer}
As we have discussed in Sec.\ref{sec:intro}, the transformer\cite{vaswani2017attention}, a sequence-to-sequence structure,  becomes a natural choice for processing 3D point clouds due to its property of permutation invariance and the ability to capture long-range dependencies. Several current works\cite{engel2021pointtransformer,pan2021pointformer,guo2021pct} do apply transformers to enhance the representation learning in 3D point clouds. \cite{guo2021pct} directly applies the attention mechanism on the entire set of points, which results in heavy memory consumption and huge computational complexity. On the contrary, \cite{zhao2021pointtransformer} groups all the points into several local regions and self-attention is adopted in each local region independently. Moreover, \cite{pan2021pointformer} first applies max-pooling on each local region, and an extra global transformer is adopted to model scene-level 
dependencies on those generated features. 

However, those methods omit the inherent relationships between neighboring local regions. Separated local regions damage the structure of instances and each local region cannot perceive the local context information around itself, which makes their features inferior for transformers to achieve discriminative representations. Although the commonly-used hierarchical architecture \cite{qi2017pointnet++,qi2019votenet,zhao2021pointtransformer} can exploit neighboring region relationships implicitly to some extent during continuous sub-sampling, we argue that it is far from being fully excavated. We propose to add connections across local regions to exchange information between them and enhance the modeling power. Thus, a novel module named Local Context Propagation (LCP) is designed and the details can be found in Sec.\ref{sec:lcp}.

The main difference between our method and other methods lies in that the interaction occurs at the point level that each point feature is updated instead of treating each local region as a whole like \cite{qi2017pointnet++, qi2019votenet, Song2022LSLPCT} do, which enhances the feature learning from a more fine-grained perspective. Extensive experiments in Sec.\ref{sec:exp} can well prove the superiority of this fine-grained design in transformers.

\begin{figure*}[htbp]
\centering
\includegraphics[width=1.0\linewidth]{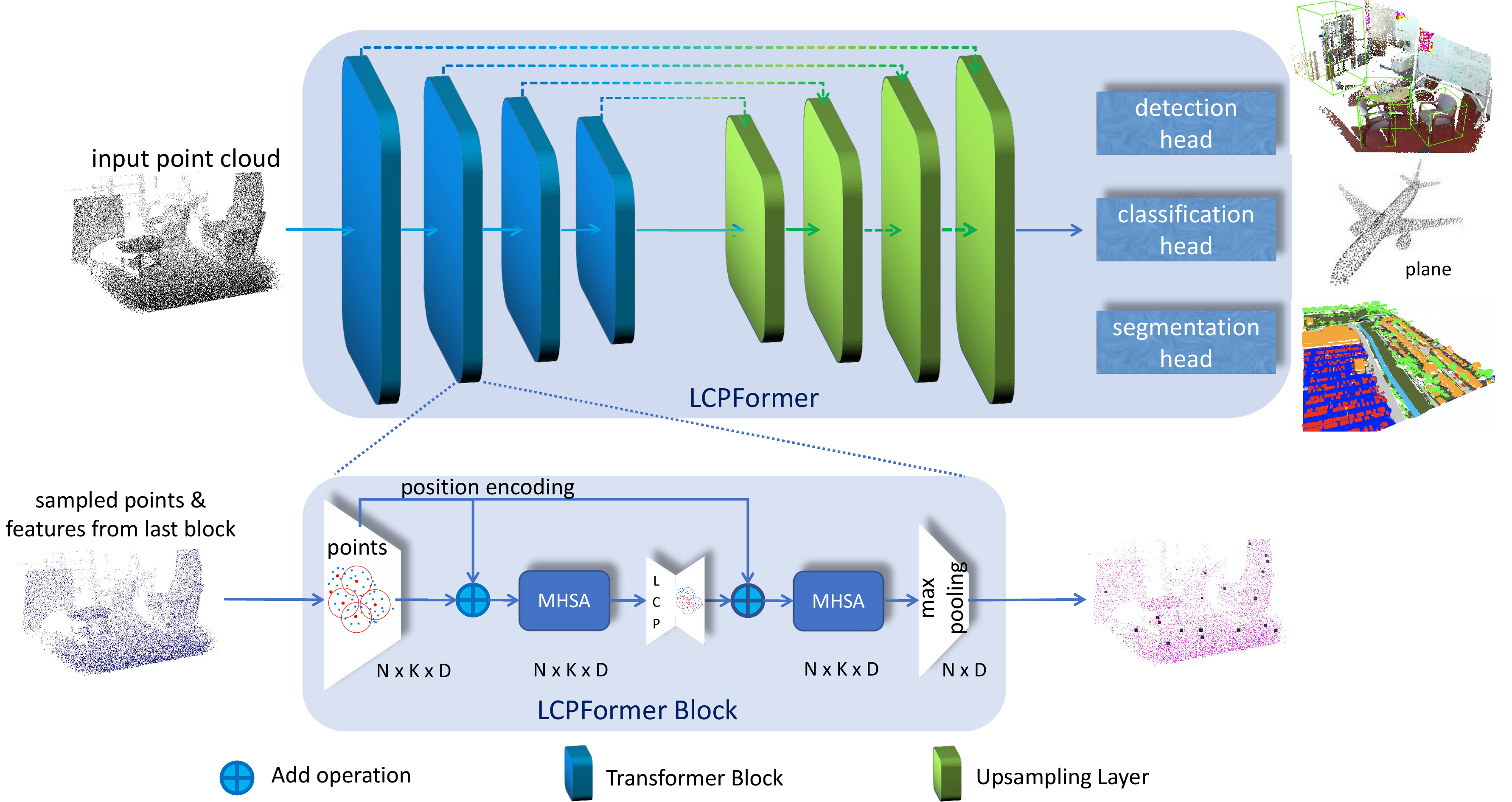}
   \caption{Overall architecture of our transformer feature extractor. The LCPFormer Block consists of a grouping layer, two self-attention layers (MHSA), a local context propagation (LCP) module, and a max-pooling layer. The backbone network is constructed by multiple LCPFormer blocks and upsampling layers. Our LCPFormer can be applied to different tasks including object detection, classification, and segmentation, according to the use of different heads.
   }
\label{overallarchi}
\end{figure*}

\subsection{Local Context Propagation}\label{sec:lcp}
\label{lcp}

As discussed in Sec.\ref{rethinkingtransformer}, when directly applying the transformer to each local region independently, it lacks sufficient information exchange among regions. There are several attempts at this issue in the image field. HRFormer\cite{yuan2021hrformer} uses a $3\times3$ depth-wise convolution to merge features from separate local windows and Swin Transformer\cite{liu2021swin} utilizes a shifted window mechanism to implicitly achieve a similar goal. However, these methods depend on the regular structure of 2D image grids and there is no effective solution for 3D point clouds due to their irregularity and disorder. To overcome this problem, we propose a simple yet effective module, named Local Context Propagation (LCP), to propagate context information between neighboring local regions on the fine-grained point level.

Our method is based on a simple observation that when dividing the whole point cloud into different local regions, naturally there is overlap among them. Take the standard settings in \cite{qi2017pointnet++,pan2021pointformer} as example. Suppose there are $N$ points in a 3D point cloud, we select ${N'} = N / 2$ center points via FPS algorithm and group neighbor points via kNN algorithm ($k = 16$ in our experiments), thus we obtain $k \times N' = 8N$ points in total. According to our statistics, most of the $N$ points are grouped multiple times repeatedly. Consequently, those \textit{shared} points can serve as suitable carriers to conduct message passing between local regions to make feature representations more informative and discriminative. Fig.\ref{motivation} (b) can well illustrate our motivation.

Briefly speaking, LCP works by updating point features in overlapping areas of different regions. As discussed above, most of the points are shared by several local regions, thus the whole point cloud features can be effectively updated. Given a point $x_i$, we denote the local regions it belongs to as $\{G_1, \dots, G_m\}$. After the transformer is independently applied to those regions, for each local region $G_j$, the point $x_i$ should have a corresponding feature in it, and we denote it as $f_i^j$. We update the feature of $x_i$ by using the weighted sum of those corresponding features:
\begin{equation}
f_i' = \sum_j w_{j} f_i^j\quad\quad j = 0, \dots, m
\label{eq:sum_weight}
\end{equation}
where $w_j$ denotes the normalized weight for region $G_j$. In this way, we successfully achieve the purpose of context information propagation between local regions.

The detailed structure of LCP can be found in Fig.\ref{cntstructure}. Assuming that the whole point cloud contains $N$ points grouped into $M$ local regions, we can denote the input of LCP as $F_{in} \in \mathbb{R}^{M \times K \times C}$, where $K$ is the number of points in each local region and $C$ is the feature dimension of each point.
To extract an effective representation for each local region, we choose to concatenate the outputs from two pooling operators: max-pooling and average-pooling on the learned point-wise features following \cite{Woo2018cbam}. The motivation behind this design is that max-pooling tends to extract the specific important features while average-pooling can involve the surrounding context. Both are important for effective point cloud analysis.
Representations $A \in \mathbb{R}^{M\times 2C}$ for $M$ local regions are obtained and sent through a $1\times1$ convolution to generate an adaptive weight matrix $W \in \mathbb{R}^{M\times C}$, which is normalized using $\mathrm{Softmax}$. 
This process can be formulated as:
\begin{equation}
  A = \mathrm{MaxPool}(F_{in}) \oplus \mathrm{AvgPool}(F_{in})
\end{equation}
\begin{equation}
  W = \mathrm{Softmax}(\mathrm{Conv}(A))
\label{eq:learnablemask}
\end{equation}
The weight matrix $W$ is applied in Eq.\ref{eq:sum_weight} to update the feature of each point and we thus obtain the final output $F_{out} \in \mathbb{R}^{N \times C}$.

The parameter number and computational complexity of the proposed LCP module are negligible as shown in the ablation study of module efficiency in Sec.\ref{lightweight}

\begin{table*}[htbp]
\caption{Performance comparisons with previous methods on SUN RGB-D\cite{song2015sun} test set.}
\centering
{\begin{tabular}{l|c|c|c|c}
    \toprule  
    Methods & Reference & Backbone & mAP@IoU0.25 & mAP@IoU0.5 \\
    \toprule
    H3DNet  \cite{zhang2020h3dnet}  & ECCV 2020 
    & 4$\times$PointNet++ & 60.1 & 39.0 \\
    MLCVNet \cite{xie2020mlcvnet} & CVPR 2020 
    & PointNet++ & 59.8 & - \\
    HGNet \cite{chen2020hgnet} & CVPR 2020 
    & GU-Net & 61.6 & - \\
    3DETR \cite{3detr2021misra} & ICCV 2021 
    & - & 58.0 & 30.3\\
    3DETR-m \cite{3detr2021misra} & ICCV 2021 
    & - & 59.1 & 32.7 \\
    T3D \cite{zhao2021transformer3ddet} & TCSVT 2021
    & PointNet++ & 60.1 & - \\
    EQ-Net \cite{yang2022qpointnet} & CVPR 2022 
    & EQ-PointNet++ & 60.5 & 38.5 \\
    \midrule
    VoteNet \cite{qi2019votenet} & ICCV 2019 
    & PointNet++ & 59.1 & 35.8 \\
    Pointformer + VoteNet \cite{pan2021pointformer} & CVPR 2021 
    & Pointformer & 61.1 & 36.9 \\
    LCPFormer+ VoteNet & N/A & w/ LCP & 61.4 & 39.6 \\
    \midrule
    Group-Free \cite{liu2021groupfree} & ICCV 2021
    & PointNet++ & 62.8 & 44.4 \\
    LCPFormer& N/A & w/o LCP & 62.7 & 43.3 \\
    \textbf{LCPFormer}& N/A & w/ LCP & \textbf{63.2} & \textbf{46.2} \\
    \bottomrule
\end{tabular}}
\label{rgbdresult}
\end{table*}

\subsection{LCPFormer}
\label{networkarchitecture}
Based on the LCP module, we propose our LCPFormer to effectively extract features from point clouds. As shown in Fig.\ref{overallarchi}, the LCPFormer is composed of multiple transformer blocks (colored in blue) and upsampling layers (colored in green). We adopt the upsampling layer in PointNet++\cite{qi2017pointnet++} and skip connections are used to enhance the upsampled features. Each LCPFormer Block comprises a grouping layer, two self-attention layers, and an LCP module in the middle.

The grouping layer groups the whole point cloud $\{f_i, x_i\}$ into different local regions $\{G_1, G_2, \dots, G_M\}$, where $M$ is the number of regions and $f_i$, $x_i$ are of size $C$ and $d$ respectively. It first samples center points via FPS and for each center $j$, a group $G_j=\{f_{i_j}, x_{i_j}\}$ of size $K\times(d+C)$ is constructed by gathering $K$ points within a local neighborhood using kNN or BallQuery\cite{qi2017pointnet++}.

The following self-attention layer performs self-attention within each local region independently to capture local features. Then, our LCP module is introduced to enhance feature interaction between neighboring local regions. Another self-attention layer is adopted for further feature optimization based on the LCP module outputs. Lastly, a common max-pooling operation is used to abstract a single $C'$-dimension representation for each local region, and the final output is in the shape of $M\times C'$.

\textbf{Network Architecture}. It is worth noting that the number of local regions, the number of self-attention layers in each block, and the number of LCPFormer Blocks may vary in different tasks. It is reasonable as different tasks share different characteristics. We construct our LCPFormer architecture utilizing the proposed LCP module and applying it to various 3D point cloud applications. For each task, we choose different hyper-parameters and slightly modify the basic architecture for better adaptation.

\subsubsection{3D Object Detection}
The backbone network for 3D object detection contains 4 LCPFormer blocks and 2 upsampling layers. The whole input point cloud is downsampled to 2048, 1024, 512, and 256 points with an increasing receptive radius of 0.2, 0.4, 0.8, and 1.2 respectively by the LCPFormer blocks, and then upsampled to 1024 points by the next two upsampling layers. In the grouping layer of each LCPFormer Block, $k=16$ points are grouped for each local region to achieve a balance between computational complexity and performance. Without loss of generality, we adopt the LCPFormer to replace the PointNet++ backbone in the original Group-Free \cite{liu2021groupfree} model and adopt the detection head structure of Group-Free to produce the final detections. To further prove the effectiveness of our method, we also adopt the LCPFormer to replace the backbone in the original VoteNet \cite{qi2019votenet} model for a fair comparison.

\subsubsection{3D Classification} 
The backbone network for shape classification is composed of 4 consecutive LCPFormer Blocks, each with one self-attention layer and one LCP module. The number of points is downsampled from 1024 and kept to 256 in all 4 blocks thus all points are selected during FPS operation. The $k$ of kNN is set to 16, 12, 8, and 8 in four blocks for multi-scale feature generation.

\begin{table}[htbp]
\caption{Performance comparisons with previous methods on ModelNet40 dataset. \textit{Note that we do not use normal vectors as the extra inputs or other special tricks as well}. * means we report the best results among public codebases and our own reproduction. \dag We build our LCPFormer on the basis of the PCT with the extra LCP module.}
\centering
\resizebox{\linewidth}{!}
{\begin{tabular}{l|c|c|c}
    \toprule  
    Methods & Reference & mAcc(\%) & OA(\%) \\
    \toprule
    3DShapeNets \cite{modelnet40dataset} & CVPR 2015 & 77.3 & 84.7 \\
    VoxNet \cite{maturana2015voxnet} & IROS 2015 & 83.0 & 85.9 \\
    Subvolume  \cite{qi2016volumetric} & CVPR 2016 & 86.0 & 89.2 \\
    PointNet   \cite{qi2017pointnet} & CVPR 2017 & 86.2 & 89.2 \\
    A-SCN  \cite{xie2018attentional} & CVPR 2018 & - & 89.8 \\
    MVCNN \cite{mvcnn2015} & ICCV 2015 & - & 90.1 \\
    SO-Net  \cite{sonet2018} & CVPR 2018 & - & 90.9 \\
    Point Transformer$_1^{*}$  \cite{zhao2021pointtransformer} & ICCV 2021 & - & 91.7 \\
    Kd-Net  \cite{kdnet2017} & ICCV 2017 & - & 91.8 \\
    PointNet++  \cite{qi2017pointnet++} & NeurIPS 2017 & - & 91.9 \\
    PointGrid  \cite{pointgrid2018le} & CVPR 2018 & - & 92.0 \\
    SpecGCN  \cite{specgcn2018} & ECCV 2018 & - & 92.1 \\
    Point-PlaneNet  \cite{pointplanenet} & DSP 2020 & 90.5 & 92.1 \\
    PCNN  \cite{pcnn2018atzmon} & TOG 2018 & - & 92.3 \\
    PointWeb  \cite{pointweb2019zhao} & CVPR 2019 & 89.4 & 92.3 \\
    SpiderCNN  \cite{spidercnn2018xu} & ECCV 2018 & - & 92.4 \\
    PointCNN  \cite{pointcnn2018li} & NeurIPS 2018 & 88.1 & 92.5 \\
    PointConv  \cite{wu2019pointconv} & CVPR 2019 & - & 92.5 \\
    A-CNN  \cite{acnn2019koma} & CVPR 2019 & - & 92.6 \\
    P2Sequence  \cite{liu2019point2sequence} & AAAI 2019 & - & 92.6 \\
    Point Transformer$_2$  \cite{engel2021pointtransformer} & Access 2021 & - & 92.8 \\
    KPConv  \cite{thomas2019kpconv} & ICCV 2019 & - & 92.9 \\
    DGCNN  \cite{wang2019dgcnn} & TOG 2019 & 90.2 & 92.9 \\
    RS-CNN  \cite{liu2019rscnn} & CVPR 2019 & - & 92.9 \\
    PointANSL   \cite{yan2020pointasnl} & CVPR 2020 & - & 92.9 \\
    InterpCNN  \cite{mao2019interpcnn} & ICCV 2019 & - & 93.0 \\
    DRNet   \cite{Qiu2021DRNet} & WACV 2021 & - & 93.1 \\
    EQ-Net  \cite{yang2022qpointnet} & CVPR 2022 & - & 93.2 \\
    PSNet \cite{Li2022psnet} & T-CSVT 2022 & - & 93.3 \\
    RSMix   \cite{lee2021rsmix} & CVPR 2021 & - & 93.5 \\
    PatchFormer  \cite{Zhang2022patchformer} & CVPR 2022 & - & 93.5 \\
    PCT (w/o LCP)\dag \cite{guo2021pct} & CVM 2021 & 90.0 & 93.2 \\
    \midrule
    \textbf{LCPFormer} & - & {\bf 90.7} & {\bf 93.6} \\
    \bottomrule
\end{tabular}}
\label{modelnet40result}
\end{table}

For better comparison, we build our backbone with similar capacity and adopt the same classification head as PCT\cite{guo2021pct}. Comparison results can be found in the ablation study of module efficiency in Sec.\ref{lightweight}.

\begin{table*}[htbp]
\caption{Performance comparisons with previous methods on S3DIS \cite{Armeni2016s3dis} and SensatUrban\cite{hu2021sensat} \textit{test} set. SensatUrban results are reported from the official leaderboard of the Urban3D ICCV21 challenge. * means we report the best results among public codebases and our own reproduction.
}
\centering
{
\begin{tabular}{lccccc} 
    \toprule
     & \multicolumn{3}{c}{S3DIS} 
     & \multicolumn{2}{c}{SensatUrban}  \\
    \cmidrule(r){2-4}
    \cmidrule(r){5-6}
    Method & 
    mIoU(\%) & mAcc(\%) & OA(\%) &
    mAcc(\%) & OA(\%) \\
    \midrule
    PointNet \cite{qi2017pointnet}  & 41.1 &23.7 & -  &23.7 & 80.8\\
    PointNet++ \cite{qi2017pointnet++} & - & - & -  & 32.9  & 84.3\\
    TragenConv \cite{TragenConv2018tangent} & 52.8 & 62.2 & 82.5  &33.3 &77.0\\
    SPGraph \cite{SPGraph2018Landrieu}  & 58.0 & 66.5 & 86.4 &37.3  &85.3 \\
    SparseConv \cite{SparseConv2018graham} & - & - & -  &42.7 &88.7\\
    LocalTransformer \cite{wang2022local} & 64.1 & 71.9 & 87.6 & - & - \\
    KPConv \cite{thomas2019kpconv} & 67.1 & 72.8 & - &57.6  &93.2\\
    RandLA-Net \cite{hu2020randla} & 62.4 & 71.4 & 87.2  &52.7 & 89.8 \\
    PatchFormer \cite{Zhang2022patchformer} & 68.1 & - & - & - & - \\
    PSNet \cite{Li2022psnet} & 62.9 & - & 87.8 & - & - \\
    DenseKPNET \cite{Li2022DenseK} & 68.9 & - & 90.8 & - & - \\
    Point Transformer$_1^{*}$ \cite{zhao2021pointtransformer} & 70.0 & 76.8 & 90.4 & - & - \\
    \midrule
    \textbf{LCPFormer w/o LCP} & \textbf{69.3} & \textbf{75.2} & \textbf{90.2} & \textbf{61.7} & \textbf{93.0} \\
    \textbf{LCPFormer w/ LCP} & \textbf{70.2} & \textbf{76.8} & \textbf{90.8} & \textbf{63.4} & \textbf{93.5}\\
    \bottomrule
\end{tabular}}
\label{sensatresult}
\end{table*}

\subsubsection{3D Semantic Segmentation}
\label{segmentationnetwork}
We build a UNet-like network\cite{ronneberger2015unet} using 4 LCPFormer blocks and 4 upsampling layers in the backbone for the semantic segmentation task since it requires point-wise features for dense prediction. A simple softmax segmentation head is adopted to output class probabilities from the extracted features. The input point number is set to 4096 and further downsampled to 1024, 512, and 256 points, respectively. We then upsample it gradually to 4096 points with four upsampling layers. The $k$ of kNN is set as 16.

\section{Experiments}
\label{sec:exp}

We apply our LCPFormer to various 3D tasks including 3D shape classification, 3D object
detection, and semantic segmentation on the most widely-used benchmarks. Please see Appendix for further benchmarks details.

\subsection{3D Object Detection}
\label{objectdetectionresult}

\textbf{SUN RGB-D} \cite{song2015sun}. We evaluate our method for 3D object detection on the commonly used SUN RGB-D dataset\cite{song2015sun}. SUN RGB-D\cite{song2015sun} is an indoor point cloud dataset for multiple scene understanding tasks. It contains $\sim$10K RGB-D images densely annotated with over 58K 3D bounding boxes in 37 different categories.

Our training settings mostly follow \cite{liu2021groupfree}. We employ an AdamW optimizer with an initial learning rate of 0.0003 and a weight decay of 0.0005. We train the network for 600 epochs and decrease the learning rate by 10$\times$ at epochs 420, 480, and 540. To augment the training data, we adopt random flips in the horizontal direction, random rotations between $[-5^{\circ}, 5^{\circ}]$, and random scaling by $[0.9, 1.1]$. Following a standard evaluation protocol, we adopt the mean Average Precision (mAP) under IoU thresholds 0.25 and 0.5 as evaluation metrics, and results of the 10 most common categories are reported.

\textbf{Results on SUN RGB-D} are shown in Tab.\ref{rgbdresult}. When adopting our LCPFormer as the backbone of VoteNet \cite{qi2019votenet}, results show that our method makes a \textbf{2.3\%} mAP@0.25 improvement and further outperform the Pointformer \cite{pan2021pointformer} by \textbf{0.3\%} mAP@0.25. Moreover, when adopting our LCPFormer as the backbone of Group-Free \cite{liu2021groupfree}, our LCPFormer achieves \textbf{63.2\%} mAP@0.25 and \textbf{46.2\%} mAP@0.5, outperforming the Group-Free by \textbf{1.8\%} mAP@0.5 and \textbf{0.4\%} mAP@0.25. The comparison between the last two rows validates the effectiveness of our LCP module and is consistent with our analysis that context propagation between local regions is helpful for accurate detection results in Fig.\ref{motivation}.

\subsection{Point Cloud Shape Classification}
\label{classificationresult}
{\bf ModelNet40} \cite{modelnet40dataset}. We benchmark our 3D shape classification network on the widely used point cloud shape classification dataset ModelNet40 \cite{modelnet40dataset}. It contains over 10,000 CAD models in 40 categories.

We follow the basic data split as well as the data augmentation in \cite{guo2021pct}. During training, a random anisotropic scaling between [$0.67, 1.5$] and a random translation between [$-0.2, 0.2$] are used successively. We set the input point number to 1024, and train the network for 250 epochs with the SGD optimizer. The initial learning rate is set to 0.0008 and decays after every epoch using the CosineAnnealing strategy. Following most of the previous works, we report two common metrics i.e. OA (Overall Accuracy) and mAcc (mean of class-wise accuracy) for evaluation.

\textbf{Results on ModelNet40} are detail in Tab.\ref{modelnet40result}. LCPFormer outperforms the PCT with \textbf{93.6\%} in OA and \textbf{90.7\%} in mAcc. \textit{Note that we do not use normal vectors as extra inputs in our experiment.}

\subsection{3D Semantic Segmentation}
\label{segmentationresult}

\textbf{SensatUrban} \cite{hu2021sensat} \& \textbf{S3DIS} \cite{Armeni2016s3dis}. We then evaluate our method for 3D semantic segmentation on both indoor and outdoor senses with two large-scale point cloud datasets. SensatUrban \cite{hu2021sensat} is a photogrammetric point cloud dataset that contains over 100 million richly annotated points. It is labeled into 13 categories, including large objects e.g. buildings and ground, and extremely small objects in an urban scene e.g. bikes and paths. S3DIS \cite{Armeni2016s3dis} is an indoor 3D RGB point clouds dataset that contains six areas of three buildings. All points are labeled with their semantic ground truth from 13 categories including board, bookcase, chair, ceiling, beam, etc.

We train the network for 100 epochs with the AdamW optimizer. The initial learning rate is set to 0.001 and decays after every epoch using the CosineAnnealing strategy. We employ the mean Intersection over Union (mIoU) as the basic evaluation metric and also report the Overall Accuracy (OA) and per-class IoU scores respectively.

\begin{table*}[htbp]
\caption{Performance comparisons with existing SOTA methods on SensatUrban\cite{hu2021sensat} \textit{test} set. Overall Accuracy (OA), mean IoU (mIoU), and per-class IoU scores are reported from the leaderboard of SensatUrban.}
\centering
{
\begin{tabular}{l|lllllllllllllll}
    \toprule  
    Methods
    & \rotatebox{55}{OA(\%)}
    & \rotatebox{55}{mIoU(\%)}
    & \rotatebox{55}{ground}
    & \rotatebox{55}{veg.}
    & \rotatebox{55}{buildings} 
    & \rotatebox{55}{walls} 
    & \rotatebox{55}{bridge} 
    & \rotatebox{55}{parking} 
    & \rotatebox{55}{rail} 
    & \rotatebox{55}{traffic.} 
    & \rotatebox{55}{street.} 
    & \rotatebox{55}{cars} 
    & \rotatebox{55}{path.} 
    & \rotatebox{55}{bikes} 
    & \rotatebox{55}{water} \\
    \toprule
    PointNet \cite{qi2017pointnet} &80.8 &23.7 &68.0 &89.5 &80.0 &0.0 &0.0 &4.0 &0.0 &31.6 &0.0 &35.1 &0.0 &0.0 &0.0\\
    PointNet++ \cite{qi2017pointnet++} &84.3 &32.9 &72.5 &94.2 &84.8 &2.7 &2.1 &25.8 &0.0 &31.5 &11.4 &38.8 &7.1 &0.0 &56.9\\
    TragenConv \cite{TragenConv2018tangent} &77.0 &33.3 &71.5 &91.4 &75.9 &35.2 &0.0 &45.3 &0.0 &26.7 &19.2 &67.6 &0.0 &0.0 &0.0\\
    SPGraph \cite{SPGraph2018Landrieu} &85.3 &37.3 &69.9 &94.6 &88.9 &32.8 &12.6 &15.8 &15.5 &30.6 &23.0 &56.4 &0.5 &0.0 &44.2\\
    SparseConv \cite{SparseConv2018graham} &88.7 &42.7 &74.1 &97.9 &94.2 &63.3 &7.5 &24.2 &0.0 &30.1 &34.0 &74.4 &0.0 &0.0 &54.8 \\
    KPConv \cite{thomas2019kpconv} &93.2 &57.6 &{\bf87.1} &{\bf98.9} &95.3 &{\bf74.4} &28.7 &41.4 &0.0 &56.0 &{\bf54.4} &{\bf85.7} &40.4 &0.0 &{\bf86.3}\\
    RandLA-Net \cite{hu2020randla} &89.8 &52.7 &80.1 &98.1 &91.6 &48.9 &40.8 &{\bf51.6} &0.0 &56.7 &33.2 &80.1 &32.6 &0.0 &71.3 \\
    \midrule
    \textbf{LCPFormer} &{\bf93.5} &{\bf63.4} &86.5 &98.3 &{\bf96.0} &55.8 &{\bf57.0} &50.6 &{\bf46.3} &{\bf61.4} &51.5 &85.2 &{\bf49.2} &0.0 &86.2\\
    \bottomrule
\end{tabular}
}
\label{fullsensatresult}
\end{table*}


\begin{figure}[htbp]
\begin{center}
    \includegraphics[width=0.9\linewidth]{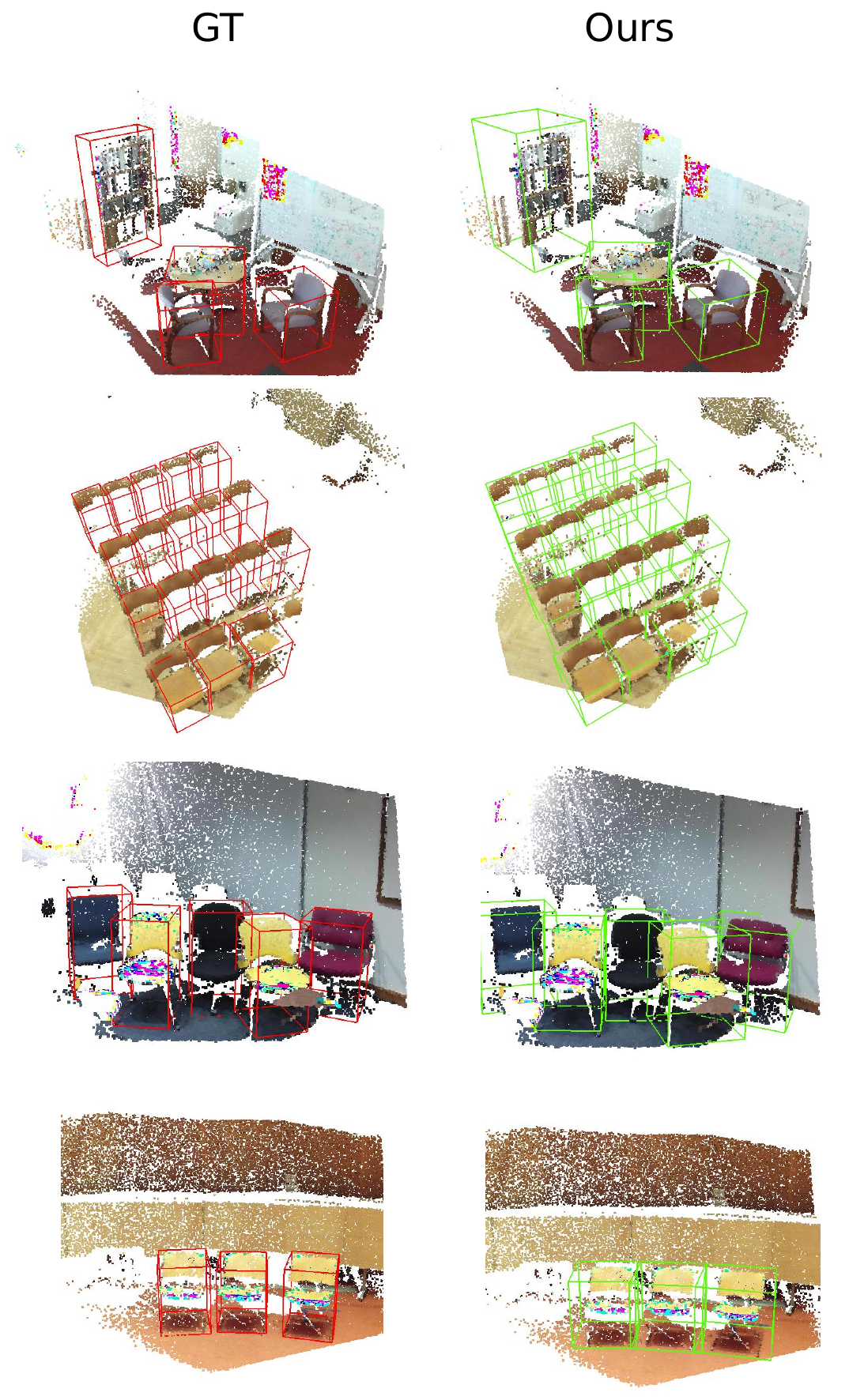}
\end{center}
   \caption{{\bf Visualization of detection results on SUN RGB-D dataset.} Left: Ground truth. Right: Our LCPFormer results.}
\label{sunrgbdviz}
\end{figure}

\begin{figure*}[htbp]
\begin{center}
    \includegraphics[width=1.0\linewidth]{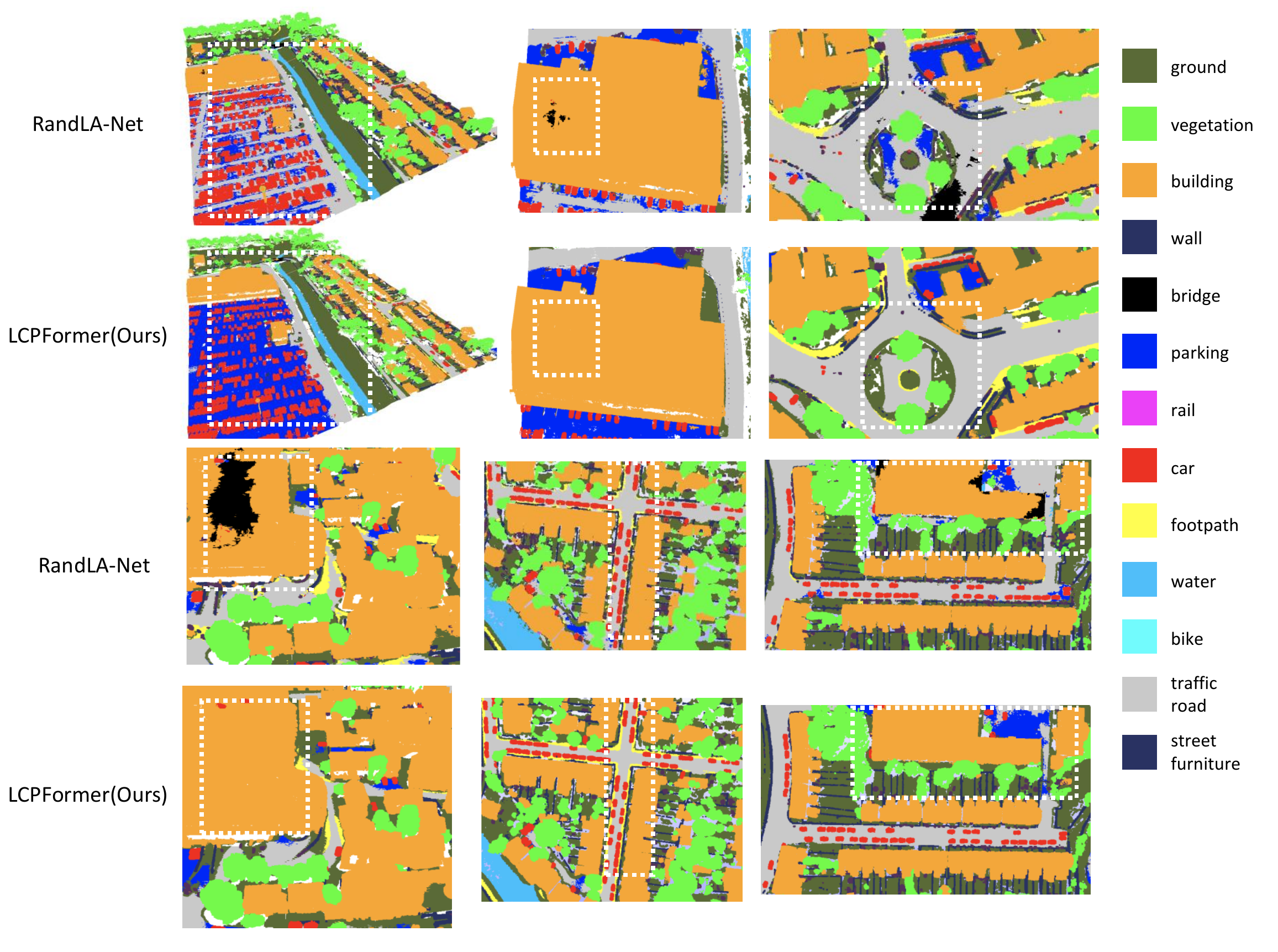}
\end{center}
   \caption{{\bf Visualization of segmentation results on SensatUrban dataset.} We randomly sample several point cloud scenes to visualize their output results. A current state-of-the-art method is selected for a more intuitive comparison. The different results of our LCPFormer and RandLA-Net are highlighted in white boxes.}
\label{sensatviz}
\end{figure*}


\textbf{Results for S3DIS} are shown in Tab.\ref{sensatresult}. Following a common protocol\cite{qi2017pointnet++}, we evaluate the presented approach with Area 5 withheld during training and used for testing. Our method achieved 70.2\% mIoU, 76.5 mAcc\% and 90.1\% OA. We also test our method on a challenging urban scale segmentation dataset SensatUrban \cite{hu2021sensat} and achieved promising results with about {\bf5.8\%} mIoU overall improvements compared with the previous best method KPConv \cite{thomas2019kpconv}. \textbf{Full results for SensatUrban} are presented in Tab.\ref{fullsensatresult}, our proposed method also has great potential when it comes to fine-grained categories. For instance, LCPFormer reaches \textbf{57.0\%} mIoU on bridge category and \textbf{46.3\%} mIoU on rail category while the scores of PointNet \cite{qi2017pointnet} remain 0.0\% mIoU on both. Similar results are also reflected in the traffic category and path category, where our method makes a \textbf{29.8\%} mIoU and \textbf{49.2\%} mIoU improvement respectively compared with PointNet \cite{qi2017pointnet}.

Our proposed LCPFormer with the LCP module remedies the information loss of instances caused by separate local regions in common transformer-based methods \cite{pan2021pointformer,zhao2021pointtransformer} via local context propagation among neighboring regions. In this way, it can achieve more informative and discriminative representations to recognize the whole structure of an instance very well, thereby reducing classification errors and contributing to its excellent performance on semantic segmentation tasks. 

\subsection{Visualization Analysis}
To evaluate the impact of our proposed LCPFormer more intuitively, we randomly sample several point cloud scenes to visualize their output results. We choose both SensatUrban and SUN RGB-D datasets for a broader comparison. The visualization results are shown in Fig.\ref{sunrgbdviz} and Fig.\ref{sensatviz}. 

Results shown in Fig.\ref{sunrgbdviz} compare our detection results with the ground-truth label. Similar outcomes can also be found in Fig.\ref{sensatviz} that strongly support our motivation, which is the separated sampling in the local region causes the destruction of the instance information and our proposed LCP can effectively improve the point cloud feature with a more informative and discriminative representation. As we can see from the second and fourth rows, some instances were incorrectly segmented. Conversely, the LCPFormer successfully maintained the integrity of the instances and classify most of the points correctly.

\subsection{Ablation Study}
\label{ablation}
In this section, we conduct extensive ablation experiments to verify our method, including the effectiveness of the proposed LCP module, and the impact of different design choices on the transformer backbone. 
All experiments are trained and evaluated on the SUN RGB-D\cite{song2015sun} dataset.

{\bf Effects of each component.} We validate the effectiveness of the proposed LCPFormer and the LCP module on the detection task. In addition to presenting the difference between our point level LCP and the natural message aggregation from the receptive field changes during downsampling, we also implement a global transformer layer to replace our LCP module just as \cite{pan2021pointformer} does for a fair comparison. Detailed results are shown in Tab.\ref{abla:comp}.

\begin{table}[htbp]
\caption{Ablation study on applying different backbones with Group-Free baseline.}
\centering
{\begin{tabular}{c|c|c|c|c}
\toprule
Backbone & LCP & GTL & mAP@IoU0.25 & mAP@IoU0.5 \\
\toprule
PointNet++ & \XSolidBrush & \XSolidBrush & 62.8 & 42.3 \\
LCPFormer  & \XSolidBrush & \XSolidBrush& 62.7 & 43.3 \\
LCPFormer  & \Checkmark & \XSolidBrush& {\bf63.2} & {\bf46.2} \\
LCPFormer  & \XSolidBrush & \Checkmark& 61.2 & 43.9\\
\bottomrule
\end{tabular}}
\label{abla:comp}
\end{table}

The first row denotes the PointNet++ baseline. We can find that if we discard the LCP module from our LCPFormer, the simple transformer-based method can achieve comparable performance with the baseline. However, if our proposed LCP module is adopted, it can accomplish a significant performance improvement, especially in the mAP@0.5 metric, which has about \textbf{3.9\%} mAP gains. This result keeps consistent with our discussion in Sec.\ref{lcp} and Sec.\ref{objectdetectionresult}. The LCP module helps transformers to achieve more accurate bounding boxes. In contrast, as shown in the last row in Tab.\ref{abla:comp}, the implemented global transformer layer (GTL) actually damages the performance to some extent as it omits the inherent relationship between neighboring local regions.

{\bf Ablation of the number of neighbor points.} Since our method highly relies on overlapping points and the number of neighbor points $k$ naturally determines the degree of overlap to a large extent. Tab.\ref{abla: knn} shows the influence of different neighbor point number $k$. Within a certain range ($k \le 16$), increasing more points to enhance the structural information in the local area brings more gains for the model performance. But when these neighboring regions go too far ($k \ge 32$), it brings excessive noise with more irrelevant points, which weakens the accuracy of the network to a certain extent. Similar results can also be found in \cite{zhao2021pointtransformer}. We adopt $k = 16$ as the default setting in our experiments for a balance between performance and computational cost.

\begin{table}[htbp]
\caption{Comparisons between different $k$ values in kNN.
}
\centering
{\begin{tabular}{c|c|c}
\toprule
$k$ & mAP@IoU0.25 & mAP@IoU0.5 \\
\toprule
4  & 56.92 & 37.53 \\
8  & 59.48 & 41.37 \\
16 & {\bf62.70} & 43.34 \\
32 & 61.52 & {\bf44.22} \\
\bottomrule
\end{tabular}}
\label{abla: knn}
\end{table}

{\bf Ablation of the transformer backbone architecture.} To deeper validate our transformer architecture, we conduct an ablation study on the number of LCPFormer blocks and the number of MHSA layers in each block. The results are summarized in Tab.\ref{abla: nblock} and Tab.\ref{abla:nlayer}. It is obvious that insufficient parameters will inevitably weaken the capacity of the network, but deeper models (block number = 5 or attention layer = 3) bring more parameters and unnecessary learning burdens, which makes the model extremely difficult to train. 

\begin{table}[htbp]
\caption{Ablation study on the number of LCPFormer blocks.} 
\centering
{\begin{tabular}{c|c|c}
\toprule
Blocks & mAP@IoU0.25 & mAP@IoU0.5 \\
\toprule
3  & 61.36 & 42.80 \\
4  & {\bf62.70} & {\bf43.34} \\
5  & 61.12 & 42.17 \\
\bottomrule
\end{tabular}}
\label{abla: nblock}
\end{table}

\begin{table}[htbp]
\caption{Ablation study on the number of attention layers. }
\centering
{\begin{tabular}{c|c|c}
\toprule
Attention Layers & mAP@IoU0.25 & mAP@IoU0.5 \\
\toprule
1  & 61.16 & 43.66 \\
2  & {\bf62.70} & 43.34 \\
3  & 62.08 & {\bf43.81} \\
\bottomrule
\end{tabular}}
\label{abla:nlayer}
\end{table}

{\bf Module efficiency analysis.}\label{lightweight}
Eq.\ref{eq:sum_weight}-\ref{eq:learnablemask} summarize our proposed LCP module.
To further validate the efficiency of our LCP module, we compare our method with two famous transformer-based methods for the 3D shape classification task and the results are shown in Tab.\ref{efficiency}. 
Common metrics including GFLOPs, model parameters, and latency are used. All the experiments are conducted on a single NVIDIA GeForce RTX 2080Ti.

\begin{table}[htbp]
\caption{Efficiency comparisons with previous methods on ModelNet40 \cite{modelnet40dataset}.}
\centering
{\begin{tabular}{c|c|c|c|c}
    \toprule
    Method & GFLOPs & Params & Latency & OA \\
    \toprule
    Point Transformer  & 147.2 & 9.58M & 18.7ms & 91.7 \\
    PCT  & 17.4 & 2.94M & 14.6ms &  93.2\\
    \textbf{LCPFormer} & 17.6 & 3.04M & 14.9ms & \textbf{93.6} \\
    \bottomrule
\end{tabular}}
\label{efficiency}
\end{table}

Note that our LCPFormer for the 3D shape classification task is designed on the basis of the PCT \cite{guo2021pct} structure as described in Sec.\ref{networkarchitecture}. The only difference between them is the extra LCP module in the LCPFormer. The comparison between PCT \cite{guo2021pct} and LCPFormer shows that our LCP module is lightweight enough yet effective. It achieves a considerable improvement of 0.4\% OA on the ModelNet40 dataset with a negligible amount of parameters and inference time.

\section{Conclusion}

This work explores the natural fit of the transformer in 3D point cloud perception and focuses on the destruction of the instance information caused by separate local regions. We present a novel and effective message exchange module named Local Context Propagation (LCP). Unlike the previous methods, our LCPFormer is tailored for irregular point clouds and enhances the inherent relationship among the neighboring local regions via local context propagation. Finally, our proposed method achieves considerable improvement compared with various transformer-based methods in multiple 3D tasks including shape classification, and dense prediction tasks such as object detection and semantic segmentation. As a sequence-to-sequence structure, transformers show great potential for sets embedded in the geometric space like point clouds. In future work, we would like to further explore the versatility of our work and implement it to even more datasets for different 3D tasks such as 3D pose estimation, 3D point cloud matching, etc.


\bibliographystyle{IEEEtran}
\bibliography{ieee}

\begin{thebibliography}{10}
\providecommand{\url}[1]{#1}
\csname url@samestyle\endcsname
\providecommand{\newblock}{\relax}
\providecommand{\bibinfo}[2]{#2}
\providecommand{\BIBentrySTDinterwordspacing}{\spaceskip=0pt\relax}
\providecommand{\BIBentryALTinterwordstretchfactor}{4}
\providecommand{\BIBentryALTinterwordspacing}{\spaceskip=\fontdimen2\font plus
\BIBentryALTinterwordstretchfactor\fontdimen3\font minus
  \fontdimen4\font\relax}
\providecommand{\BIBforeignlanguage}[2]{{%
\expandafter\ifx\csname l@#1\endcsname\relax
\typeout{** WARNING: IEEEtran.bst: No hyphenation pattern has been}%
\typeout{** loaded for the language `#1'. Using the pattern for}%
\typeout{** the default language instead.}%
\else
\language=\csname l@#1\endcsname
\fi
#2}}
\providecommand{\BIBdecl}{\relax}
\BIBdecl

\bibitem{deng2021h2rcnn}
J.~Deng, W.~Zhou, Y.~Zhang, and H.~Li, ``From multi-view to hollow-3d:
  Hallucinated hollow-3d r-cnn for 3d object detection,'' \emph{IEEE
  Transactions on Circuits and Systems for Video Technology}, vol.~31, no.~12,
  pp. 4722--4734, 2021.

\bibitem{Liu2021PQANet}
Q.~Liu, H.~Yuan, H.~Su, H.~Liu, Y.~Wang, H.~Yang, and J.~Hou, ``Pqa-net: Deep
  no reference point cloud quality assessment via multi-view projection,''
  \emph{IEEE Transactions on Circuits and Systems for Video Technology},
  vol.~31, no.~12, pp. 4645--4660, 2021.

\bibitem{yang2018pixor}
B.~Yang, W.~Luo, and R.~Urtasun, ``Pixor: Real-time 3d object detection from
  point clouds,'' in \emph{Proceedings of the IEEE conference on Computer
  Vision and Pattern Recognition}, 2018, pp. 7652--7660.

\bibitem{liang2018deep}
M.~Liang, B.~Yang, S.~Wang, and R.~Urtasun, ``Deep continuous fusion for
  multi-sensor 3d object detection,'' in \emph{Proceedings of the European
  Conference on Computer Vision (ECCV)}, 2018, pp. 641--656.

\bibitem{ku2018joint}
J.~Ku, M.~Mozifian, J.~Lee, A.~Harakeh, and S.~L. Waslander, ``Joint 3d
  proposal generation and object detection from view aggregation,'' in
  \emph{2018 IEEE/RSJ International Conference on Intelligent Robots and
  Systems (IROS)}.\hskip 1em plus 0.5em minus 0.4em\relax IEEE, 2018, pp. 1--8.

\bibitem{yan2018second}
Y.~Yan, Y.~Mao, and B.~Li, ``Second: Sparsely embedded convolutional
  detection,'' \emph{Sensors}, vol.~18, no.~10, p. 3337, 2018.

\bibitem{song2016deep}
S.~Song and J.~Xiao, ``Deep sliding shapes for amodal 3d object detection in
  rgb-d images,'' in \emph{Proceedings of the IEEE conference on computer
  vision and pattern recognition}, 2016, pp. 808--816.

\bibitem{SparseConv2018graham}
B.~Graham, M.~Engelcke, L.~Van Der~Maaten, and xx, ``3d semantic segmentation
  with submanifold sparse convolutional networks,'' in \emph{Proceedings of the
  IEEE conference on computer vision and pattern recognition}, 2018, pp.
  9224--9232.

\bibitem{Minkowski}
C.~Choy, J.~Gwak, and S.~Savarese, ``4d spatio-temporal convnets: Minkowski
  convolutional neural networks,'' in \emph{Proceedings of the IEEE/CVF
  Conference on Computer Vision and Pattern Recognition}, 2019, pp. 3075--3084.

\bibitem{lang2019pointpillars}
A.~H. Lang, S.~Vora, H.~Caesar, L.~Zhou, J.~Yang, and O.~Beijbom,
  ``Pointpillars: Fast encoders for object detection from point clouds,'' in
  \emph{Proceedings of the IEEE/CVF Conference on Computer Vision and Pattern
  Recognition}, 2019, pp. 12\,697--12\,705.

\bibitem{3dsis}
J.~Hou, A.~Dai, and M.~Nie{\ss}ner, ``3d-sis: 3d semantic instance segmentation
  of rgb-d scans,'' in \emph{Proceedings of the IEEE/CVF Conference on Computer
  Vision and Pattern Recognition}, 2019, pp. 4421--4430.

\bibitem{pillar}
Y.~Wang, A.~Fathi, A.~Kundu, D.~A. Ross, C.~Pantofaru, T.~Funkhouser, and
  J.~Solomon, ``Pillar-based object detection for autonomous driving,'' in
  \emph{Computer Vision--ECCV 2020: 16th European Conference, Glasgow, UK,
  August 23--28, 2020, Proceedings, Part XXII 16}.\hskip 1em plus 0.5em minus
  0.4em\relax Springer, 2020, pp. 18--34.

\bibitem{zhou2018voxelnet}
Y.~Zhou and O.~Tuzel, ``Voxelnet: End-to-end learning for point cloud based 3d
  object detection,'' in \emph{Proceedings of the IEEE conference on computer
  vision and pattern recognition}, 2018, pp. 4490--4499.

\bibitem{qi2017pointnet}
C.~R. Qi, H.~Su, K.~Mo, and L.~J. Guibas, ``Pointnet: Deep learning on point
  sets for 3d classification and segmentation,'' in \emph{Proceedings of the
  IEEE conference on computer vision and pattern recognition}, 2017, pp.
  652--660.

\bibitem{qi2017pointnet++}
C.~Qi, L.~Yi, H.~Su, and L.~J. Guibas, ``Pointnet++: Deep hierarchical feature
  learning on point sets in a metric space,'' in \emph{NIPS}, 2017.

\bibitem{thomas2019kpconv}
H.~Thomas, C.~R. Qi, J.-E. Deschaud, B.~Marcotegui, F.~Goulette, and L.~J.
  Guibas, ``Kpconv: Flexible and deformable convolution for point clouds,'' in
  \emph{Proceedings of the IEEE/CVF International Conference on Computer
  Vision}, 2019, pp. 6411--6420.

\bibitem{hu2020randla}
Q.~Hu, B.~Yang, L.~Xie, S.~Rosa, Y.~Guo, Z.~Wang, N.~Trigoni, and A.~Markham,
  ``Randla-net: Efficient semantic segmentation of large-scale point clouds,''
  in \emph{Proceedings of the IEEE/CVF Conference on Computer Vision and
  Pattern Recognition}, 2020, pp. 11\,108--11\,117.

\bibitem{qi2019votenet}
C.~R. Qi, O.~Litany, K.~He, and L.~J. Guibas, ``Deep hough voting for 3d object
  detection in point clouds,'' in \emph{Proceedings of the IEEE/CVF
  International Conference on Computer Vision}, 2019, pp. 9277--9286.

\bibitem{yogo}
C.~Xu, B.~Zhai, B.~Wu, T.~Li, W.~Zhan, P.~Vajda, K.~Keutzer, and M.~Tomizuka,
  ``You only group once: Efficient point-cloud processing with token
  representation and relation inference module,'' \emph{arXiv preprint
  arXiv:2103.09975}, 2021.

\bibitem{guo2021pct}
M.-H. Guo, J.-X. Cai, Z.-N. Liu, T.-J. Mu, R.~R. Martin, and S.-M. Hu, ``Pct:
  Point cloud transformer,'' \emph{Computational Visual Media}, vol.~7, no.~2,
  pp. 187--199, 2021.

\bibitem{pan2021pointformer}
X.~Pan, Z.~Xia, S.~Song, L.~E. Li, and G.~Huang, ``3d object detection with
  pointformer,'' in \emph{Proceedings of the IEEE/CVF Conference on Computer
  Vision and Pattern Recognition}, 2021, pp. 7463--7472.

\bibitem{zhao2021pointtransformer}
H.~Zhao, L.~Jiang, J.~Jia, P.~H. Torr, and V.~Koltun, ``Point transformer,'' in
  \emph{Proceedings of the IEEE/CVF International Conference on Computer
  Vision}, 2021, pp. 16\,259--16\,268.

\bibitem{engel2021pointtransformer}
N.~Engel, V.~Belagiannis, K.~Dietmayer, and xx, ``Point transformer,''
  \emph{IEEE Access}, 2021.

\bibitem{liu2021groupfree}
Z.~Liu, Z.~Zhang, Y.~Cao, H.~Hu, and X.~Tong, ``Group-free 3d object detection
  via transformers,'' \emph{arXiv preprint arXiv:2104.00678}, 2021.

\bibitem{wang2021spatial}
J.~Wang, R.~Chakraborty, and X.~Y. Stella, ``Spatial transformer for 3d point
  clouds,'' \emph{IEEE Transactions on Pattern Analysis and Machine
  Intelligence}, 2021.

\bibitem{zhu2020deformableDETR}
X.~Zhu, W.~Su, L.~Lu, B.~Li, X.~Wang, and J.~Dai, ``Deformable detr: Deformable
  transformers for end-to-end object detection,'' \emph{arXiv preprint
  arXiv:2010.04159}, 2020.

\bibitem{carion2020DETR}
N.~Carion, F.~Massa, G.~Synnaeve, N.~Usunier, A.~Kirillov, and S.~Zagoruyko,
  ``End-to-end object detection with transformers,'' in \emph{European
  Conference on Computer Vision}.\hskip 1em plus 0.5em minus 0.4em\relax
  Springer, 2020, pp. 213--229.

\bibitem{dosovitskiy2020ViT}
A.~Dosovitskiy, L.~Beyer, A.~Kolesnikov, D.~Weissenborn, X.~Zhai,
  T.~Unterthiner, M.~Dehghani, M.~Minderer, G.~Heigold, S.~Gelly \emph{et~al.},
  ``An image is worth 16x16 words: Transformers for image recognition at
  scale,'' \emph{arXiv preprint arXiv:2010.11929}, 2020.

\bibitem{chen2021crossvit}
C.-F. Chen, Q.~Fan, and R.~Panda, ``Crossvit: Cross-attention multi-scale
  vision transformer for image classification,'' \emph{arXiv preprint
  arXiv:2103.14899}, 2021.

\bibitem{d2021convit}
S.~d'Ascoli, H.~Touvron, M.~Leavitt, A.~Morcos, G.~Biroli, and L.~Sagun,
  ``Convit: Improving vision transformers with soft convolutional inductive
  biases,'' \emph{arXiv preprint arXiv:2103.10697}, 2021.

\bibitem{yuan2021hrformer}
Y.~Yuan, R.~Fu, L.~Huang, W.~Lin, C.~Zhang, X.~Chen, and J.~Wang, ``Hrformer:
  High-resolution transformer for dense prediction,'' \emph{arXiv preprint
  arXiv:2110.09408}, 2021.

\bibitem{han2021transformer}
K.~Han, A.~Xiao, E.~Wu, J.~Guo, C.~Xu, and Y.~Wang, ``Transformer in
  transformer,'' \emph{arXiv preprint arXiv:2103.00112}, 2021.

\bibitem{liu2021swin}
Z.~Liu, Y.~Lin, Y.~Cao, H.~Hu, Y.~Wei, Z.~Zhang, S.~Lin, and B.~Guo, ``Swin
  transformer: Hierarchical vision transformer using shifted windows,''
  \emph{arXiv preprint arXiv:2103.14030}, 2021.

\bibitem{TragenConv2018tangent}
M.~Tatarchenko, J.~Park, V.~Koltun, and Q.-Y. Zhou, ``Tangent convolutions for
  dense prediction in 3d,'' in \emph{Proceedings of the IEEE Conference on
  Computer Vision and Pattern Recognition}, 2018, pp. 3887--3896.

\bibitem{mvcnn2015}
H.~Su, S.~Maji, E.~Kalogerakis, and E.~Learned-Miller, ``Multi-view
  convolutional neural networks for 3d shape recognition,'' in
  \emph{Proceedings of the IEEE international conference on computer vision},
  2015, pp. 945--953.

\bibitem{qi2016volumetric}
C.~R. Qi, H.~Su, M.~Nie{\ss}ner, A.~Dai, M.~Yan, and L.~J. Guibas, ``Volumetric
  and multi-view cnns for object classification on 3d data,'' in
  \emph{Proceedings of the IEEE conference on computer vision and pattern
  recognition}, 2016, pp. 5648--5656.

\bibitem{kanezaki2018rotationnet}
A.~Kanezaki, Y.~Matsushita, and Y.~Nishida, ``Rotationnet: Joint object
  categorization and pose estimation using multiviews from unsupervised
  viewpoints,'' in \emph{Proceedings of the IEEE conference on computer vision
  and pattern recognition}, 2018, pp. 5010--5019.

\bibitem{maturana2015voxnet}
D.~Maturana and S.~Scherer, ``Voxnet: A 3d convolutional neural network for
  real-time object recognition,'' in \emph{2015 IEEE/RSJ International
  Conference on Intelligent Robots and Systems (IROS)}.\hskip 1em plus 0.5em
  minus 0.4em\relax IEEE, 2015, pp. 922--928.

\bibitem{song2017semantic}
S.~Song, F.~Yu, A.~Zeng, A.~X. Chang, M.~Savva, and T.~Funkhouser, ``Semantic
  scene completion from a single depth image,'' in \emph{Proceedings of the
  IEEE conference on computer vision and pattern recognition}, 2017, pp.
  1746--1754.

\bibitem{wu2019pointconv}
W.~Wu, Z.~Qi, and L.~Fuxin, ``Pointconv: Deep convolutional networks on 3d
  point clouds,'' in \emph{Proceedings of the IEEE/CVF Conference on Computer
  Vision and Pattern Recognition}, 2019, pp. 9621--9630.

\bibitem{yang2019gumbelsampling}
J.~Yang, Q.~Zhang, B.~Ni, L.~Li, J.~Liu, M.~Zhou, and Q.~Tian, ``Modeling point
  clouds with self-attention and gumbel subset sampling,'' in \emph{Proceedings
  of the IEEE/CVF conference on computer vision and pattern recognition}, 2019,
  pp. 3323--3332.

\bibitem{dovrat2019snet}
O.~Dovrat, I.~Lang, and S.~Avidan, ``Learning to sample,'' in \emph{Proceedings
  of the IEEE/CVF Conference on Computer Vision and Pattern Recognition}, 2019,
  pp. 2760--2769.

\bibitem{zhao2021transformer3ddet}
L.~Zhao, J.~Guo, D.~Xu, and L.~Sheng, ``Transformer3d-det: Improving 3d object
  detection by vote refinement,'' \emph{IEEE Transactions on Circuits and
  Systems for Video Technology}, vol.~31, no.~12, pp. 4735--4746, 2021.

\bibitem{Li2022psnet}
L.~Li, L.~He, J.~Gao, and X.~Han, ``Psnet: Fast data structuring for
  hierarchical deep learning on point cloud,'' \emph{IEEE Transactions on
  Circuits and Systems for Video Technology}, vol.~32, no.~10, pp. 6835--6849,
  2022.

\bibitem{guan2021m3detr}
T.~Guan, J.~Wang, S.~Lan, R.~Chandra, Z.~Wu, L.~Davis, and D.~Manocha,
  ``M3detr: Multi-representation, multi-scale, mutual-relation 3d object
  detection with transformers,'' \emph{arXiv preprint arXiv:2104.11896}, 2021.

\bibitem{ye2020hvnet}
M.~Ye, S.~Xu, and T.~Cao, ``Hvnet: Hybrid voxel network for lidar based 3d
  object detection,'' in \emph{Proceedings of the IEEE/CVF conference on
  computer vision and pattern recognition}, 2020, pp. 1631--1640.

\bibitem{shi2020pv}
S.~Shi, C.~Guo, L.~Jiang, Z.~Wang, J.~Shi, X.~Wang, and H.~Li, ``Pv-rcnn:
  Point-voxel feature set abstraction for 3d object detection,'' in
  \emph{Proceedings of the IEEE/CVF Conference on Computer Vision and Pattern
  Recognition}, 2020, pp. 10\,529--10\,538.

\bibitem{pointgrid2018le}
T.~Le and Y.~Duan, ``Pointgrid: A deep network for 3d shape understanding,'' in
  \emph{Proceedings of the IEEE conference on computer vision and pattern
  recognition}, 2018, pp. 9204--9214.

\bibitem{SPGraph2018Landrieu}
L.~Landrieu, M.~Simonovsky, xx, and xx, ``Large-scale point cloud semantic
  segmentation with superpoint graphs,'' in \emph{Proceedings of the IEEE
  conference on computer vision and pattern recognition}, 2018, pp. 4558--4567.

\bibitem{shen2018kcnet}
Y.~Shen, C.~Feng, Y.~Yang, and D.~Tian, ``Mining point cloud local structures
  by kernel correlation and graph pooling,'' in \emph{Proceedings of the IEEE
  conference on computer vision and pattern recognition}, 2018, pp. 4548--4557.

\bibitem{wang2019graphattn}
L.~Wang, Y.~Huang, Y.~Hou, S.~Zhang, and J.~Shan, ``Graph attention convolution
  for point cloud semantic segmentation,'' in \emph{Proceedings of the IEEE/CVF
  conference on computer vision and pattern recognition}, 2019, pp.
  10\,296--10\,305.

\bibitem{Li2019deepgcn}
G.~Li, M.~Muller, A.~Thabet, and B.~Ghanem, ``Deepgcns: Can gcns go as deep as
  cnns?'' in \emph{Proceedings of the IEEE/CVF International Conference on
  Computer Vision (ICCV)}, October 2019.

\bibitem{wang2019dgcnn}
Y.~Wang, Y.~Sun, Z.~Liu, S.~E. Sarma, M.~M. Bronstein, and J.~M. Solomon,
  ``Dynamic graph cnn for learning on point clouds,'' \emph{Acm Transactions On
  Graphics (tog)}, vol.~38, no.~5, pp. 1--12, 2019.

\bibitem{pointweb2019zhao}
H.~Zhao, L.~Jiang, C.-W. Fu, and J.~Jia, ``Pointweb: Enhancing local
  neighborhood features for point cloud processing,'' in \emph{Proceedings of
  the IEEE/CVF Conference on Computer Vision and Pattern Recognition}, 2019,
  pp. 5565--5573.

\bibitem{touvron2021going}
H.~Touvron, M.~Cord, A.~Sablayrolles, G.~Synnaeve, and H.~J{\'e}gou, ``Going
  deeper with image transformers,'' \emph{arXiv preprint arXiv:2103.17239},
  2021.

\bibitem{wu2021cvt}
H.~Wu, B.~Xiao, N.~Codella, M.~Liu, X.~Dai, L.~Yuan, and L.~Zhang, ``Cvt:
  Introducing convolutions to vision transformers,'' \emph{arXiv preprint
  arXiv:2103.15808}, 2021.

\bibitem{graham2021levit}
B.~Graham, A.~El-Nouby, H.~Touvron, P.~Stock, A.~Joulin, H.~J{\'e}gou, and
  M.~Douze, ``Levit: a vision transformer in convnet's clothing for faster
  inference,'' \emph{arXiv preprint arXiv:2104.01136}, 2021.

\bibitem{yuan2021tokens}
L.~Yuan, Y.~Chen, T.~Wang, W.~Yu, Y.~Shi, Z.~Jiang, F.~E. Tay, J.~Feng, and
  S.~Yan, ``Tokens-to-token vit: Training vision transformers from scratch on
  imagenet,'' \emph{arXiv preprint arXiv:2101.11986}, 2021.

\bibitem{jiang2021all}
Z.~Jiang, Q.~Hou, L.~Yuan, D.~Zhou, Y.~Shi, X.~Jin, A.~Wang, and J.~Feng, ``All
  tokens matter: Token labeling for training better vision transformers,''
  \emph{arXiv preprint arXiv:2104.10858}, 2021.

\bibitem{srinivas2021bottleneck}
A.~Srinivas, T.-Y. Lin, N.~Parmar, J.~Shlens, P.~Abbeel, and A.~Vaswani,
  ``Bottleneck transformers for visual recognition,'' in \emph{Proceedings of
  the IEEE/CVF Conference on Computer Vision and Pattern Recognition}, 2021,
  pp. 16\,519--16\,529.

\bibitem{huang2019interlaced}
L.~Huang, Y.~Yuan, J.~Guo, C.~Zhang, X.~Chen, and J.~Wang, ``Interlaced sparse
  self-attention for semantic segmentation,'' \emph{arXiv preprint
  arXiv:1907.12273}, 2019.

\bibitem{vaswani2021scaling}
A.~Vaswani, P.~Ramachandran, A.~Srinivas, N.~Parmar, B.~Hechtman, and
  J.~Shlens, ``Scaling local self-attention for parameter efficient visual
  backbones,'' in \emph{Proceedings of the IEEE/CVF Conference on Computer
  Vision and Pattern Recognition}, 2021, pp. 12\,894--12\,904.

\bibitem{wang2020hrnet}
J.~Wang, K.~Sun, T.~Cheng, B.~Jiang, C.~Deng, Y.~Zhao, D.~Liu, Y.~Mu, M.~Tan,
  X.~Wang \emph{et~al.}, ``Deep high-resolution representation learning for
  visual recognition,'' \emph{IEEE transactions on pattern analysis and machine
  intelligence}, 2020.

\bibitem{vaswani2017attention}
A.~Vaswani, N.~Shazeer, N.~Parmar, J.~Uszkoreit, L.~Jones, A.~N. Gomez,
  {\L}.~Kaiser, and I.~Polosukhin, ``Attention is all you need,'' in
  \emph{Advances in neural information processing systems}, 2017, pp.
  5998--6008.

\bibitem{zhao2020exploring}
H.~Zhao, J.~Jia, and V.~Koltun, ``Exploring self-attention for image
  recognition,'' in \emph{Proceedings of the IEEE/CVF Conference on Computer
  Vision and Pattern Recognition}, 2020, pp. 10\,076--10\,085.

\bibitem{ronneberger2015unet}
O.~Ronneberger, P.~Fischer, and T.~Brox, ``U-net: Convolutional networks for
  biomedical image segmentation,'' in \emph{International Conference on Medical
  image computing and computer-assisted intervention}.\hskip 1em plus 0.5em
  minus 0.4em\relax Springer, 2015, pp. 234--241.

\bibitem{qiu2022putransformer}
S.~Qiu, S.~Anwar, and N.~Barnes, ``Pu-transformer: Point cloud upsampling
  transformer,'' in \emph{Proceedings of the Asian Conference on Computer
  Vision}, 2022, pp. 2475--2493.

\bibitem{Song2022LSLPCT}
Y.~Song, F.~He, Y.~Duan, T.~Si, and J.~Bai, ``Lslpct: An enhanced local
  semantic learning transformer for 3-d point cloud analysis,'' \emph{IEEE
  Transactions on Geoscience and Remote Sensing}, vol.~60, pp. 1--13, 2022.

\bibitem{Woo2018cbam}
S.~Woo, J.~Park, J.-Y. Lee, and I.~S. Kweon, ``Cbam: Convolutional block
  attention module,'' in \emph{Proceedings of the European Conference on
  Computer Vision (ECCV)}, September 2018.

\bibitem{song2015sun}
S.~Song, S.~P. Lichtenberg, and J.~Xiao, ``Sun rgb-d: A rgb-d scene
  understanding benchmark suite,'' in \emph{Proceedings of the IEEE conference
  on computer vision and pattern recognition}, 2015, pp. 567--576.

\bibitem{zhang2020h3dnet}
Z.~Zhang, B.~Sun, H.~Yang, and Q.~Huang, ``H3dnet: 3d object detection using
  hybrid geometric primitives,'' in \emph{European Conference on Computer
  Vision}.\hskip 1em plus 0.5em minus 0.4em\relax Springer, 2020, pp. 311--329.

\bibitem{xie2020mlcvnet}
Q.~Xie, Y.-K. Lai, J.~Wu, Z.~Wang, Y.~Zhang, K.~Xu, and J.~Wang, ``Mlcvnet:
  Multi-level context votenet for 3d object detection,'' in \emph{Proceedings
  of the IEEE/CVF conference on computer vision and pattern recognition}, 2020,
  pp. 10\,447--10\,456.

\bibitem{chen2020hgnet}
J.~Chen, B.~Lei, Q.~Song, H.~Ying, D.~Z. Chen, and J.~Wu, ``A hierarchical
  graph network for 3d object detection on point clouds,'' in \emph{Proceedings
  of the IEEE/CVF conference on computer vision and pattern recognition}, 2020,
  pp. 392--401.

\bibitem{3detr2021misra}
I.~Misra, R.~Girdhar, and A.~Joulin, ``An end-to-end transformer model for 3d
  object detection,'' in \emph{Proceedings of the IEEE/CVF International
  Conference on Computer Vision}, 2021, pp. 2906--2917.

\bibitem{yang2022qpointnet}
Z.~Yang, L.~Jiang, Y.~Sun, B.~Schiele, and J.~Jia, ``A unified query-based
  paradigm for point cloud understanding,'' 2022.

\bibitem{modelnet40dataset}
Z.~Wu, S.~Song, A.~Khosla, F.~Yu, L.~Zhang, X.~Tang, and J.~Xiao, ``3d
  shapenets: A deep representation for volumetric shapes,'' in
  \emph{Proceedings of the IEEE conference on computer vision and pattern
  recognition}, 2015, pp. 1912--1920.

\bibitem{xie2018attentional}
S.~Xie, S.~Liu, Z.~Chen, and Z.~Tu, ``Attentional shapecontextnet for point
  cloud recognition,'' in \emph{Proceedings of the IEEE Conference on Computer
  Vision and Pattern Recognition}, 2018, pp. 4606--4615.

\bibitem{sonet2018}
J.~Li, B.~M. Chen, and G.~H. Lee, ``So-net: Self-organizing network for point
  cloud analysis,'' in \emph{Proceedings of the IEEE conference on computer
  vision and pattern recognition}, 2018, pp. 9397--9406.

\bibitem{kdnet2017}
R.~Klokov and V.~Lempitsky, ``Escape from cells: Deep kd-networks for the
  recognition of 3d point cloud models,'' in \emph{Proceedings of the IEEE
  International Conference on Computer Vision}, 2017, pp. 863--872.

\bibitem{specgcn2018}
C.~Wang, B.~Samari, K.~Siddiqi, and xx, ``Local spectral graph convolution for
  point set feature learning,'' in \emph{Proceedings of the European conference
  on computer vision (ECCV)}, 2018, pp. 52--66.

\bibitem{pointplanenet}
S.~M. Peyghambarzadeh, F.~Azizmalayeri, H.~Khotanlou, and A.~Salarpour,
  ``Point-planenet: Plane kernel based convolutional neural network for point
  clouds analysis,'' \emph{Digital Signal Processing}, vol.~98, p. 102633,
  2020.

\bibitem{pcnn2018atzmon}
M.~Atzmon, H.~Maron, Y.~Lipman, and xx, ``Point convolutional neural networks
  by extension operators,'' \emph{ACM Transactions on Graphics (TOG)}, vol.~37,
  pp. 1 -- 12, 2018.

\bibitem{spidercnn2018xu}
Y.~Xu, T.~Fan, M.~Xu, L.~Zeng, and Y.~Qiao, ``Spidercnn: Deep learning on point
  sets with parameterized convolutional filters,'' in \emph{Proceedings of the
  European Conference on Computer Vision (ECCV)}, 2018, pp. 87--102.

\bibitem{pointcnn2018li}
Y.~Li, R.~Bu, M.~Sun, W.~Wu, X.~Di, and B.~Chen, ``Pointcnn: Convolution on
  x-transformed points,'' \emph{Advances in neural information processing
  systems}, vol.~31, pp. 820--830, 2018.

\bibitem{acnn2019koma}
A.~Komarichev, Z.~Zhong, J.~Hua, and xx, ``A-cnn: Annularly convolutional
  neural networks on point clouds,'' in \emph{Proceedings of the IEEE/CVF
  Conference on Computer Vision and Pattern Recognition}, 2019, pp. 7421--7430.

\bibitem{liu2019point2sequence}
X.~Liu, Z.~Han, Y.-S. Liu, and M.~Zwicker, ``Point2sequence: Learning the shape
  representation of 3d point clouds with an attention-based sequence to
  sequence network,'' in \emph{Proceedings of the AAAI Conference on Artificial
  Intelligence}, vol.~33, no.~01, 2019, pp. 8778--8785.

\bibitem{liu2019rscnn}
Y.~Liu, B.~Fan, S.~Xiang, and C.~Pan, ``Relation-shape convolutional neural
  network for point cloud analysis,'' in \emph{Proceedings of the IEEE/CVF
  Conference on Computer Vision and Pattern Recognition}, 2019, pp. 8895--8904.

\bibitem{yan2020pointasnl}
X.~Yan, C.~Zheng, Z.~Li, S.~Wang, and S.~Cui, ``Pointasnl: Robust point clouds
  processing using nonlocal neural networks with adaptive sampling,'' in
  \emph{Proceedings of the IEEE/CVF Conference on Computer Vision and Pattern
  Recognition}, 2020, pp. 5589--5598.

\bibitem{mao2019interpcnn}
J.~Mao, X.~Wang, and H.~Li, ``Interpolated convolutional networks for 3d point
  cloud understanding,'' in \emph{Proceedings of the IEEE/CVF International
  Conference on Computer Vision}, 2019, pp. 1578--1587.

\bibitem{Qiu2021DRNet}
S.~Qiu, S.~Anwar, and N.~Barnes, ``Dense-resolution network for point cloud
  classification and segmentation,'' \emph{2021 IEEE Winter Conference on
  Applications of Computer Vision (WACV)}, pp. 3812--3821, 2021.

\bibitem{lee2021rsmix}
D.~Lee, J.~Lee, J.~Lee, H.~Lee, M.~Lee, S.~Woo, and S.~Lee, ``Regularization
  strategy for point cloud via rigidly mixed sample,'' in \emph{Proceedings of
  the IEEE/CVF Conference on Computer Vision and Pattern Recognition}, 2021,
  pp. 15\,900--15\,909.

\bibitem{Zhang2022patchformer}
C.~Zhang, H.~Wan, X.~Shen, and Z.~Wu, ``Patchformer: An efficient point
  transformer with patch attention,'' in \emph{Proceedings of the IEEE/CVF
  Conference on Computer Vision and Pattern Recognition (CVPR)}, June 2022, pp.
  11\,799--11\,808.

\bibitem{Armeni2016s3dis}
I.~Armeni, O.~Sener, A.~R. Zamir, H.~Jiang, I.~Brilakis, M.~Fischer, and
  S.~Savarese, ``3d semantic parsing of large-scale indoor spaces,'' in
  \emph{Proceedings of the IEEE Conference on Computer Vision and Pattern
  Recognition (CVPR)}, June 2016.

\bibitem{hu2021sensat}
Q.~Hu, B.~Yang, S.~Khalid, W.~Xiao, N.~Trigoni, and A.~Markham, ``Towards
  semantic segmentation of urban-scale 3d point clouds: A dataset, benchmarks
  and challenges,'' in \emph{Proceedings of the IEEE/CVF Conference on Computer
  Vision and Pattern Recognition}, 2021, pp. 4977--4987.

\bibitem{wang2022local}
Z.~Wang, Y.~Wang, L.~An, J.~Liu, and H.~Liu, ``Local transformer network on 3d
  point cloud semantic segmentation,'' \emph{Information}, vol.~13, no.~4, p.
  198, 2022.

\bibitem{Li2022DenseK}
Y.~Li, X.~Li, Z.~Zhang, F.~Shuang, Q.~Lin, and J.~Jiang, ``Densekpnet: Dense
  kernel point convolutional neural networks for point cloud semantic
  segmentation,'' \emph{IEEE Transactions on Geoscience and Remote Sensing},
  vol.~60, pp. 1--13, 2022.

\end{thebibliography}


\vspace{11pt}

\begin{IEEEbiographynophoto}{Zhuoxu Huang}
received a B.E. degree in Geodesy and Geomatics Engineering from Wuhan University, Wuhan, China, in 2021. He is currently pursuing a Ph.D. degree with the Department of Computer Science, Aberystwyth University, Aberystwyth, U.K. His research interests include video understanding, 3D vision, and artificial intelligence.
\end{IEEEbiographynophoto}

\begin{IEEEbiographynophoto}{Zhiyou Zhao}
received a B.E. degree in Computer Science and Technology from Shanghai Jiao Tong University, Shanghai, China, in 2022. He is currently pursuing a M.S. degree in Data Science in the Department of Data Science, Chinese University of Hong Kong, Shenzhen. His research interests include 3D point cloud perception, object detection and artificial intelligence.

\end{IEEEbiographynophoto}

\begin{IEEEbiographynophoto}{Banghuai Li}
received a B.E. degree in the School of Software from Wuhan University, Wuhan, China, in 2015 and an M.S. degree in the School of Electronics Engineering and Computer Science from Peking University, Beijing, China, in 2018. He is currently a Staff R\&D Engineer in Momenta, a leading autonomous driving technology company in China. . Before that, he was a Researcher at Megvii Company in China, where he finished this work. His research interests include 2D/3D vision, autonomous driving, and artificial intelligence.
\end{IEEEbiographynophoto}

\begin{IEEEbiographynophoto}{Jungong Han}
is Chair Professor in Computer Vision at the Department of Computer Science, University of Sheffield, U.K. He holds the Honorary Professorship with Aberystwyth University and the University of Warwick, U.K. His research interests include computer vision, artificial intelligence, and machine learning.
\end{IEEEbiographynophoto}

\end{document}